\newif\ifisTR

\isTRtrue

\documentclass{article} 

\usepackage{fullpage}

\usepackage{microtype}
\usepackage{graphicx}
\usepackage{booktabs} 
\usepackage{placeins}
\usepackage{algorithm}
\usepackage{algorithmic}
\usepackage{subcaption}
\usepackage{tcolorbox}
\usepackage{nicefrac}  
\usepackage{enumitem}
\usepackage{hyperref}
\usepackage{multirow}
\usepackage{colortbl}
\usepackage{xspace}
\usepackage{wrapfig}
\usepackage{amsmath}
\usepackage{amssymb}
\usepackage{mathtools}
\usepackage[export]{adjustbox}
\usepackage{amsthm}
\usepackage{xcolor}
\usepackage{tikz}

\usepackage[capitalize,noabbrev]{cleveref}

\usepackage{overpic}

\theoremstyle{plain}

\theoremstyle{definition}

\theoremstyle{remark}

\definecolor{mygray}{gray}{0.85}
\definecolor{LightBlue}{cmyk}{0.06, 0.03, 0.01, 0.0}

\usepackage[textsize=tiny]{todonotes}

\usepackage[round,semicolon,sort]{natbib}

\definecolor{LightBlue}{HTML}{F3F3F3}

\newcommand{\ourmethod}{\texttt{TempBalance}\xspace}

\newcommand{\ALPHAHILL}{\texttt{PL\_Alpha\_Hill}\xspace}
\newcommand{\SPECTRALNORM}{\texttt{Spectral\_Norm}\xspace}
\newcommand{\STABLERANK}{\texttt{Stable\_Rank}\xspace}
\newcommand{\LINEARMAP}{\texttt{TB\_Linear\_Map}\xspace}
\newcommand{\SIGMOID}{\texttt{TB\_Sigmoid}\xspace}

\renewcommand{\cite}[1]{\citep{#1}}
\title{Model Balancing Helps Low-data Training and Fine-tuning}
\date{}

\author{
 Zihang Liu$^*$\textsuperscript{1, 2},
 Yuanzhe Hu$^*$\textsuperscript{1, 3},
 Tianyu Pang\textsuperscript{1},
 Yefan Zhou\textsuperscript{1},
\\
 Pu Ren\textsuperscript{4},
 Yaoqing Yang\textsuperscript{1}
\\
 \textsuperscript{1}Dartmouth College \\
 \textsuperscript{2}University of California, Berkeley \\
 \textsuperscript{3}University of California, San Diego \\
 \textsuperscript{4}Lawrence Berkeley National Lab \\
}

\begin{document}
\maketitle

\def\thefootnote{*}\footnotetext{Equal contribution. Work completed during an internship at Dartmouth College.}

\renewcommand{\thefootnote}{\arabic{footnote}}
\begin{abstract}

Recent advances in foundation models have emphasized the need to align pre-trained models with specialized domains using small, curated datasets. Studies on these foundation models underscore the importance of low-data training and fine-tuning. This topic, well-known in natural language processing (NLP), has also gained increasing attention in the emerging field of scientific machine learning (SciML). To address the limitations of low-data training and fine-tuning, we draw inspiration from Heavy-Tailed Self-Regularization (HT-SR) theory, analyzing the shape of empirical spectral densities (ESDs) and revealing an imbalance in training quality across different model layers. To mitigate this issue, we adapt a recently proposed layer-wise learning rate scheduler, \ourmethod, which effectively balances training quality across layers and enhances low-data training and fine-tuning for both NLP and SciML tasks. Notably, \ourmethod demonstrates increasing performance gains as the amount of available tuning data decreases. Comparative analyses further highlight the effectiveness of \ourmethod and its adaptability as an ``add-on'' method for improving model performance.

\end{abstract}
\section{Introduction}
Recent surges in foundation models (FMs) have stimulated research on aligning pre-trained models with specialized domains using small-sized datasets. This ``pre-train and fine-tune'' paradigm is prevalent in natural language processing (NLP) tasks~\citep{wang2019glue, wang2020superglue, rajpurkar-etal-2016-squad, lu2022learn}. It is also gaining popularity in other machine learning (ML) fields, such as scientific machine learning (SciML)~\citep{subramanian2024towards, lanusse2023astroclip, mccabe2023multiple, wu2023towards,hao2024dpot,chen2024data}.
From a practical perspective, the challenge of fine-tuning often lies in curating high-quality datasets (possibly with labeled examples) to achieve alignment with the new domain.
In SciML, people often use FMs for training on different types of partial differential equations (PDEs)~\citep{mccabe2023multiple, wu2023towards,hao2024dpot} and fine-tuning it on a certain domain when accessible scientific data from that domain is limited. As a concrete example, turbulence simulations at extremely high Reynolds numbers are computationally intensive and time-consuming, often leading to only a few available trajectories. Therefore, training SciML FMs on trajectories with different Reynolds numbers and fine-tuning it on trajectories simulated at extremely high ones is beneficial for solving the problem of poor training performance caused by insufficient data volume. 
Using SciML FMs, researchers can train these models to generalize across a wider range of downstream tasks, thereby enhancing their applicability and efficiency in diverse scientific scenarios. Prior research has shown that strong performance can indeed be achieved by fine-tuning with a few carefully selected examples~\citep{zhou2023lima}, but training with low data can still lead to unstable performance~\citep{zhang2021revisiting}. Therefore, finding fine-tuning algorithms that improve performance in low-data settings, especially few-shot alignment, becomes crucial.

\begin{figure*}[htb!]
    \centering
    \begin{subfigure}{0.99\textwidth}
    \includegraphics[width=\textwidth,keepaspectratio]{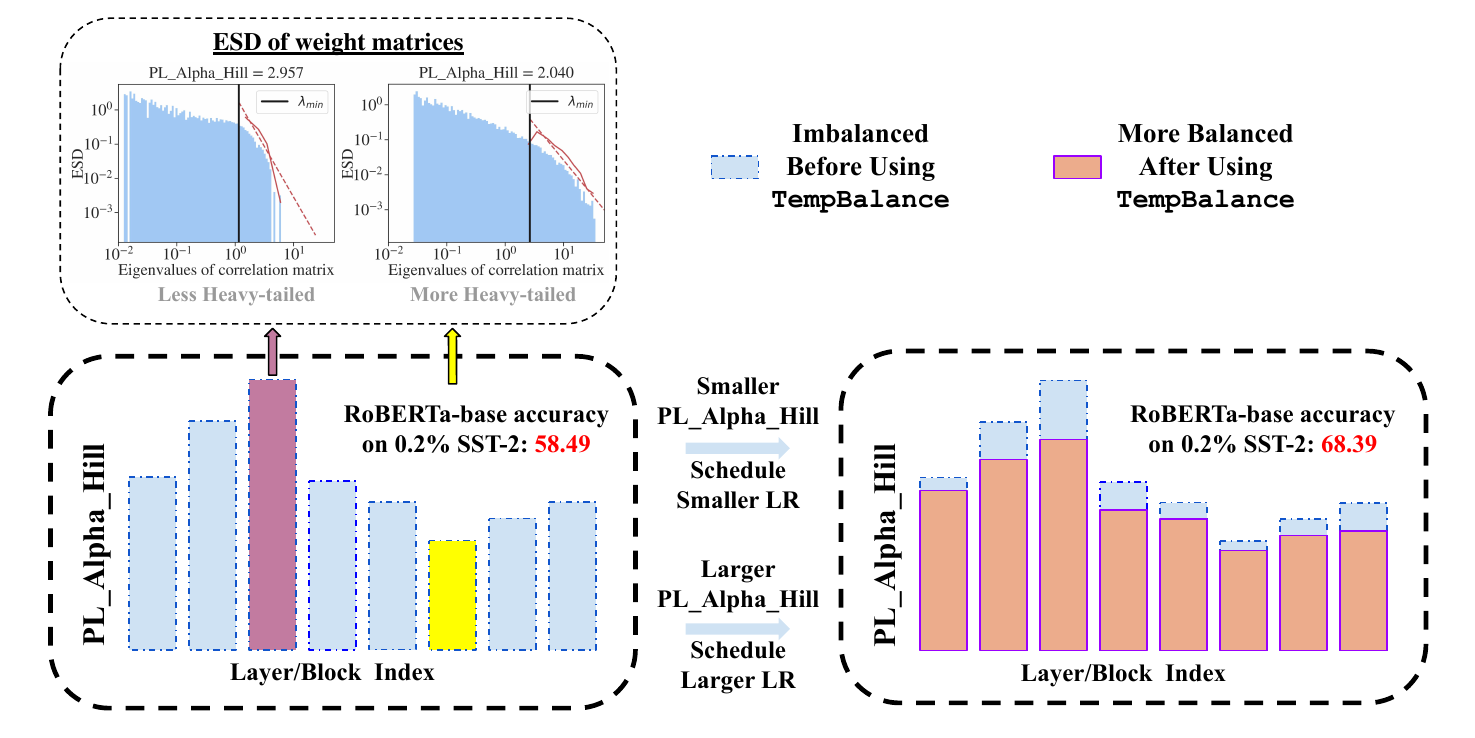} 
    \end{subfigure}  
    \caption{
    \textbf{Heavy-tail ESD analysis and \ourmethod learning rate schedule.} To characterize the heavy-tailed structure of ESD, we fit a power-law exponent \ALPHAHILL on the tail part of the ESDs (blue histograms at top left), shown as the red dashed line on the histogram. Given the imbalanced layer-wise \ALPHAHILL (bottom left), \ourmethod assigns lower learning rate to layers with lower \ALPHAHILL (more heavy-tailed), and assign higher learning rate to layers with higher \ALPHAHILL (less heavy-tailed). \ourmethod aims to balance the \ALPHAHILL distribution across layers in low-data regimes (bottom right).\looseness-1
    }
    \label{fig:ESDs}
\end{figure*}

In this work, we draw inspiration from Heavy-Tailed Self-Regularization (HT-SR) theory~\citep{martin2021implicit, martin2021predicting}, to improve model performance in low-data regimes. HT-SR theory proposes that well-trained neural network (NN) models exhibit strong correlations in weights, resulting in a Heavy-Tail (HT) structure in the Empirical Spectral Density (ESD, usually represented by a histogram of eigenvalue distribution) of each layers' weight matrix. To quantify the HT structure, we can fit a power law (PL) distribution to the HT part of the ESD and extract its exponent, namely \ALPHAHILL (see Figure~\ref{fig:ESDs}). HT-SR theory suggests that a more HT ESD (lower \ALPHAHILL) represents better training quality, and vice versa. This estimation of model and layer quality has been shown to be effective in recent work on model selection~\citep{martin2021predicting, martin2020heavy, martin2022postmortem, yang2023test}, layer-wise hyperparameter tuning~\citep{zhou2024temperature}, and pruning of large language models (LLMs)~\citep{lu2024alpha}.

Using HT-SR theory, we analyze the limitations of model training in low-data regimes by measuring the layer-wise distribution of \ALPHAHILL (discussed in~\ref{sec:diagnosis}). Our main finding is that when we train with sufficient data, \ALPHAHILL  becomes more evenly distributed across layers, resulting in better layer-wise balance; in this case, high performance can be achieved without layer-specific manipulations. However, when we reduce the number of training data samples, test performance decreases, and the standard deviation (STD) of \ALPHAHILL across layers tends to increase (see Figure~\ref{fig:trend_plot}), indicating that \ALPHAHILL is more unevenly distributed when training with fewer data, resulting in worse layer-wise balance. This finding indicates that different layers' training quality becomes more poorly aligned as we reduce training data. Therefore, layer-wise balancing is beneficial to balance under-trained layers and over-trained layers in low data regimes.

Motivated by this observation, we incorporate the variance of \ALPHAHILL across layers with the recently proposed layer-wise learning rate scheduling algorithm \ourmethod~\citep{zhou2024temperature}, to design a novel method to balance the training quality across layers.
To evaluate its empirical performance, we use \ourmethod in curated low-data regime in LLM fine-tuning and SciML tasks. We compare \ourmethod with commonly used baseline methods and demonstrate that \ourmethod not only achieves superior performance in low-data training and fine-tuning, but also can be used as a plug-in method on top of existing optimization methods to achieve even better test performance and stability, such as SAM~\citep{foret2021sharpnessaware} and AdaFactor~\citep{shazeer2018adafactor}.
Furthermore, in our analysis, we reveal that \ourmethod successfully balances training quality across all layers during training from the HT-SR point of view. We show that \ourmethod balances the training quality of each layer by reducing the STD of \ALPHAHILL of all layers. We summarize our contributions as follows \footnote{In order that our results can be reproduced and extended, we have open-sourced our code.
\href{https://github.com/ZihangHLiu/ModelBalancing}{https://github.com/ZihangHLiu/ModelBalancing.}}:

\begin{itemize}
    \item We find that low-data fine-tuning is a crucial training paradigm that can lead to imbalanced training quality across different layers of the model, measured by the large STD of \ALPHAHILL values across layers. 
    \item We focus on low-data training scenarios and demonstrate the effectiveness of using \ourmethod to balance layers and improve the performance of both NLP and SciML models. For example, we show that \ourmethod can improve RoBERTa-base trained on SST2 dataset by at most 9.9\% and increase the test accuracy of LLaMA-7B on ScienceQA dataset by at most 1.97\%, and reduce the normalized root-mean-squared-error (nRMSE) of FNO trained on 2D Compressible Navier-Stokes(CFD)\footnote{CFD means compressible fluid dynamics or, equivalently, the compressible Navier-Stokes equations.} dataset by 14.47\%. Furthermore, we show that \ourmethod achieves gradually increased performance gains as the number of data points decreases.
    \item In LM fine-tuning tasks, we demonstrate that \ourmethod can achieve better fine-tuning performance compared to baselines (including SAM~\citep{foret2021sharpnessaware} and AdaFactor~\citep{shazeer2018adafactor}) and can be used as an add-on method to combine with these existing optimization methods to achieve further improvements.
\end{itemize}

\section{Related Work}
\paragraph{Heavy-tailed Phenomenon.}
Recently, several studies have observed that a well-trained deep NN exhibits HT spectra in its weight matrices. Many papers focus on investigating the cause of the emergence of HT, and they have attributed HT spectra (or limiting HT distributions of weights) to strong correlation in weight elements~\citep{martin2021implicit,martin2021predicting}, feature learning~\citep{wang2024spectral,kothapalli2024crafting}, the Kesten–Goldie mechanism~\citep{hodgkinson2021multiplicative, gurbuzbalaban2021heavy}, $\alpha$-stable Lévy process~\citep{gurbuzbalaban2021heavy, simsekli2020hausdorff}, and the maximum-entropy principle~\citep{xie2024overlooked}.
More importantly, several studies have shown that the heavytailness of the weight spectra is strongly correlated with the quality of neural networks. For example, \citet{martin2021implicit} proposed HT-SR theory, demonstrating that the degree of HT in the ESD of each layer can be used to predict model quality: the heavier the tail of the ESD, the better the quality of the model. In addition, \citet{simsekli2020hausdorff, hodgkinson2022generalization, wang2024near} proved generalization bounds dependent on the HT distributions in either model weights or the ESDs of the weight matrices, which are validated through extensive experiments.
Motivated by these studies, some efforts have begun to leverage the degree of HT for model training~\citep{zhou2024temperature,li2024owlore,Qing2024alphaexpert}, model selection~\citep{agrawal2022alpha,yang2023test}, and model compression~\citep{barsbey2021heavy,lu2024alpha}, as well as to enhance model robustness~\citep{nassar20201}.

\paragraph{Resource-constrained Fine-tuning.}
The pretraining and fine-tuning paradigm has been a primary method for adapting foundation models to downstream tasks for resource-limited users. When adapting very large models, people often resort to the Low-Rank Adaptation method (LoRA)~\citep{hu2021lora}, which is also considered in this paper. Our primary focus is on low-data fine-tuning, an increasingly studied paradigm where the emphasis is often on careful data selection~\citep{zhou2023lima}. Furthermore, when training models in a few-shot fashion, such as \emph{in-context learning}~\citep{brown2020language, logan2021cutting, zhang-etal-2022-active}, data selection plays a crucial role in improving model performance. Our paper, however, explores layer-balancing schemes to improve model performance.

\paragraph{Data-constrainted Training and Fine-tuning in SciML.}
There has been an increasing interest in the use of ML methods to solve scientific problems~\citep{raissi2019physics,li2020fourier,karniadakis2021physics,wang2023scientific}.

One representative line of work is on neural operators~\citep{li2020fourier,lu2021learning,hao2023gnot,raonic2024convolutional}. 
These operators have demonstrated their effectiveness in scientific modeling. However, they require extensive scientific datasets. Generating high-fidelity numerical datasets is computationally demanding. Hence, to mitigate the costs associated with simulation, self-supervised pretraining has been introduced for operator learning~\citep{chen2024data}. Additionally, in low-data regimes, researchers also propose to incorporate physical laws into ML models to facilitate the learning of the underlying governing equations, often through soft regularization constraints~\citep{raissi2019physics}. Nevertheless, the physics-constrained ML strategy is limited to specific PDE scenarios (e.g., fixed PDE coefficients)~\citep{ye2024pdeformer}, which poses challenges to generalization.

\section{Methodology}
In this section, we first revisit HT-SR theory and important HT-SR metrics related to model performance. Then, we discuss \ourmethod~\cite{zhou2024temperature}, which works well on different model architectures based on ``shape metrics'' from HT-SR Theory.

\subsection{HT-SR Theory}
HT-SR theory~\citep{martin2021implicit} demonstrates the empirical fact that very well-trained models tend to exhibit strong correlations in weights, resulting in HT structure in the ESD of each layer.
Its underlying motivation stems from random matrix theory and statistical physics, as well as the observation that HT ESDs are ubiquitous in well-trained NN models.

\paragraph{Obtaining the ESD of Weight Matrices. }
To obtain the ESDs of a model, we take an NN with $L$ layers and the corresponding weight matrices $\mathbf{W}_1,\mathbf{W}_2,\cdots ,\mathbf{W}_L$.
For the $i$-th layer, 
we calculate the eigenvalues of its correlation matrix $\mathbf{X}_i=\mathbf{W}_i^\top \mathbf{W}_i$.
Then, we plot the ESD for that layer, which is the empirical distribution of these eigenvalues. 
During training, the ESD will typically gradually change to have an HT structure.
There are many metrics that have been proposed to study the properties of ESDs, among which shape metrics (metrics that depict the shape of ESD) have been shown to predict the training quality of each layer~\citep{yang2023test}. 

\paragraph{Analyzing ESDs with PL Fitting. }
To obtain robust shape metrics that predict layer quality, we fit a PL distribution to the heavy-tailed part of the ESD within an interval ($\lambda_\text{min}$, $\lambda_\text{max}$). The PL fit has the following formula:
\begin{equation}\label{eqn:ALPHA}
    p(\lambda) \propto \lambda^{-\alpha}, \lambda_\text{min} < \lambda < \lambda_\text{max}.
\end{equation}
We then extract its exponent $\alpha$ as an empirical metric. To fit a PL distribution to the ESD, we use the Hill Estimator~\citep{1975hill, zhou2024temperature}: for the $i$-th layer, suppose the weight matrix is $\mathbf{W}_i$ and the correlation matrix $\mathbf{W}_i^\top\mathbf{W}_i$ has ascending eigenvalues $\lbrace \lambda_{i} \rbrace_{i=1}^n$. 
The Hill estimator calculates \ALPHAHILL as: 
\begin{equation}\label{eqn:hill_estimator}
\ALPHAHILL= 1+\frac{k}{(\sum_{i=1}^k \ln\frac{\lambda_{n-i+1}}{\lambda_{n-k}})}.
\end{equation}
where $k$ is an adjustable parameter.

\paragraph{\ALPHAHILL Distribution and Model Quality. }When using \ALPHAHILL to analyze model performance, related works suggest that a layer with smaller \ALPHAHILL tends to be relatively ``overtrained'' (compared to other layers in the model), while layers with higher \ALPHAHILL are relatively ``undertrained.'' \citep{zhou2024temperature} find that in CV tasks, models trained with optimized hyperparameter scheduling outperform baseline methods and yield a more concentrated \ALPHAHILL distribution across layers, suggesting that a more uniformly distributed \ALPHAHILL has more balanced training quality across layers, leading to better overall quality of the model.

\subsection{\ourmethod Algorithm}\label{sec:tb_algorithm}
Prior research~\citep{martin2021implicit} has shown that temperature-like parameters significantly influence the HT structure of individual layers' ESDs. Therefore, to balance the shape of ESDs across layers, we propose to adapt the \ourmethod algorithm, which dynamically tunes the learning rate on a layer-wise basis, as the learning rate is the most important temperature parameter. Smaller learning rates are assigned to layers with more heavy-tailed ESDs to slow down the training, while larger learning rates are assigned to those with more light-tailed ESDs to accelerate the training.
We propose a novel method to map the \ALPHAHILL of each layer to the layer-wise learning rate. We first calculate their difference with the mean \ALPHAHILL value across all layers, then rescale the difference using a sigmoid-like function. Finally, we use the rescaled value as the exponent to assign the new learning rate $f_t(i)$ for the layer. We refer to this scheduling algorithm as \SIGMOID. The equations are as follows:

\begin{equation}\label{eqn:outer_eq}
f_t(i)= \eta_t \cdot 10^\phi,
\end{equation}
\begin{equation}\label{eqn:inner_eq}
  \phi = s \cdot \left(\frac{1}{1 + e^{-\tau \cdot (\alpha_i - \overline{\alpha})}} - 0.5\right),
\end{equation}
where $\eta_t$ is the base learning rate at step $t$, $\alpha_i$ is the \ALPHAHILL of layer $i$, and $\overline{\alpha}$ is the mean \ALPHAHILL across all layers. Note that $s$ and $\tau$ are tunable hyperparameters in experiments, and we often obtain the best results when we set $\tau = 10$. In \ourmethod, if a layer’s  \ALPHAHILL is higher than the mean, a learning rate higher than the base learning rate is assigned, and if it is lower, a lower learning rate is assigned. Furthermore, layers with \ALPHAHILL significantly different from the mean receive more substantial adjustments, while those closer to the mean receive minimal changes.
The intuition of this scheduling function is that it not only controls \ALPHAHILL by adjusting the learning rate based on its value, but also takes the difference of \ALPHAHILL to the mean into account to reduce the variance of \ALPHAHILL across layers by assigning learning rate changes proportional to the difference, finally balancing the training quality. In Table~\ref{tb:tb_func_comparison}, we empirically show that \SIGMOID works better than other layer-wise learning rate scheduling methods.\looseness-1
  
\paragraph{Using \ourmethod on Transformers.}
For Transformer-based architectures, we note each Transformer block consists of different types of layers (such as Query, Output, and Down Projection) with different matrix sizes, resulting in distinct ESD shapes. Therefore, we explore a more favorable scheduling method to eliminate unfair comparison of \ALPHAHILL of different ESD shapes. We reschedule each blocks' learning rate by averaging the \ALPHAHILL across all layers within the block, while in each block we use the same learning rate across all layers. In Table~\ref{tb:layer_block_comp} in Appendix~\ref{appendix:schedule_ablation}, we show that the per-block scheduling method consistently outperforms the per-layer method in different low-data regimes. Given such a design, we note that a ``layer'' used in this work when discussing Transformer-based models refers to a Transformer block.

\begin{figure*}[!htb]
    \centering
    \begin{subfigure}{0.6\linewidth}
        \includegraphics[width=\textwidth]{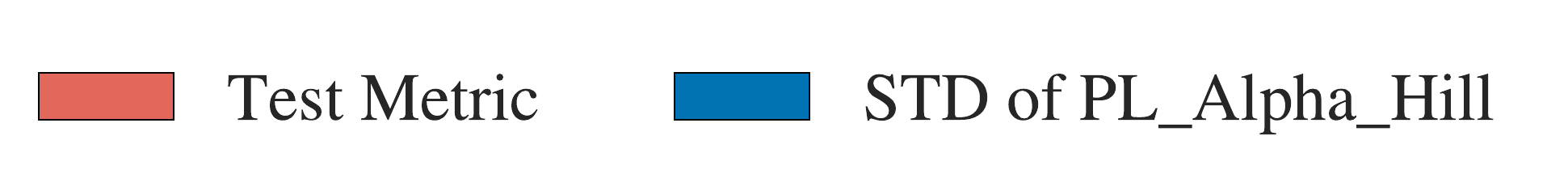}
        \label{fig:test_results_trend}
    \end{subfigure}
    \begin{subfigure}[t]{0.45\linewidth}

        \begin{overpic}[width=\textwidth]{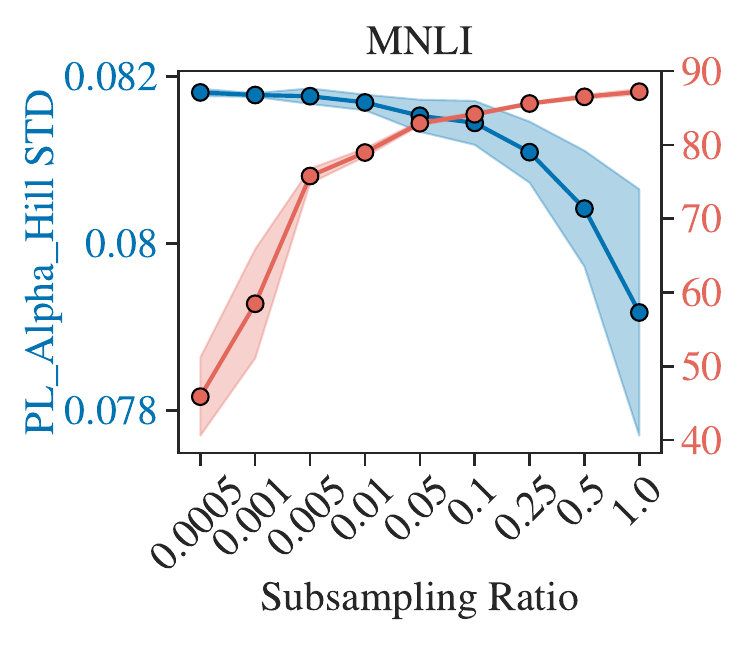}
        \put(33,28){\includegraphics[width=0.5\textwidth]{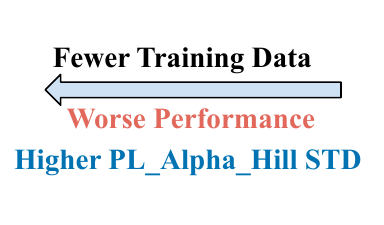}}
        \end{overpic}
    
        \label{fig:test_results_trend_mnli}
    \end{subfigure}
    \centering
    \begin{subfigure}[t]{0.45\linewidth}
        \begin{overpic}[width=\textwidth]{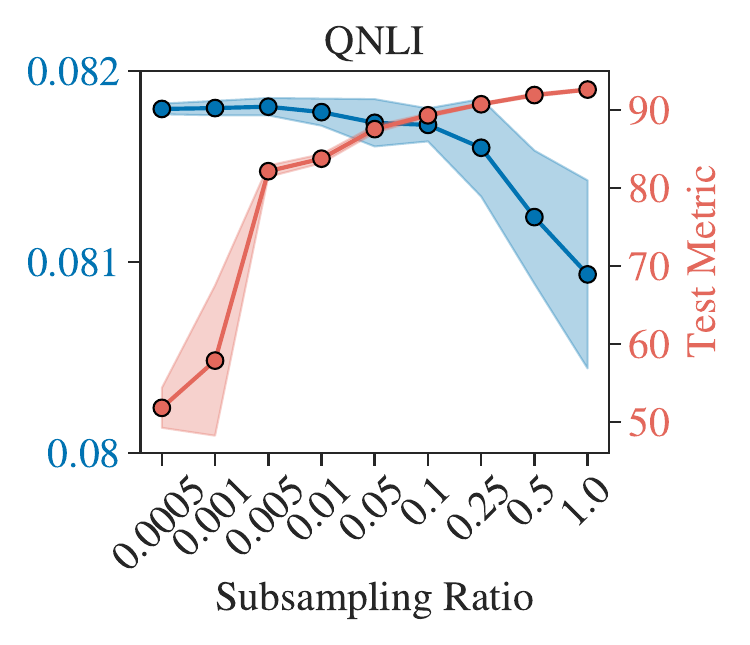}
        \put(29,28){\includegraphics[width=0.48\textwidth]{figures/predicting_trends/trends.pdf}}
        \end{overpic}

        \label{fig:alpha_std_trend_qnli}
    \end{subfigure}

    \caption{Test performance and STD of \ALPHAHILL across all layers of RoBERTa-base model trained on MNLI (Accuracy$\uparrow$) and QNLI (Accuracy$\uparrow$) under different subsampling ratios. 
    }
    \label{fig:trend_plot} 
\end{figure*}

\section{Empirical Results}

In this section, we employ HT metrics to diagnose model performance in data-limited regimes and demonstrate the effectiveness of \ourmethod in addressing data limitation in two fields: NLP and SciML. In Section~\ref{sxn:experimental_setup}, we describe our experimental setup. In Section~\ref{sec:diagnosis}, we study the correlation between ESD behaviors and model performance with limited training data. Then, in Section~\ref{sxn:low_data_results}, we evaluate \ourmethod in our experimental setup. In Section~\ref{sxn:comparison}, we compare our methods with other optimization baselines. We analyze the experimental results in Section~\ref{sec:analysis}. Finally, we perform ablation studies in Section~\ref{sxn:ablation_study}.

\subsection{Experimental Setup}\label{sxn:experimental_setup}
\noindent \textbf{Models and Evaluation.} For NLP, we evaluate \ourmethod with two widely-used fine-tuning methods: Full fine-tuning (FT) and LoRA fine-tuning~\citep{hu2021lora} using the Huggingface framework~\citep{wolf2020huggingfaces}.
We select two models with distinct sizes: RoBERTa-base~\citep{liu2019roberta} and LLaMA2-7b~\citep{touvron2023llama}. We train the models on subsampled common fine-tuning datasets, including GLUE~\citep{wang2019glue}, SuperGLUE~\citep{wang2020superglue}, SQuAD~\citep{rajpurkar-etal-2016-squad}, and ScienceQA~\citep{lu2022learn}. We train with sampling ratios ranging from 0.02\% to 50\% to evaluate our method. We also evaluate \ourmethod on low-resource datasets from three specialized domains: BioMed, CS, and News.
We choose five datasets from these domains: RCT with 500 samples~\citep{dernoncourt2017pubmed}, SciCite~\citep{Cohan2019Structural}, ChemProt~\citep{kringelum2016chemprot}, SciERC~\citep{luan2018multi}, and Hyperpartisan News~\citep{kiesel2019semeval}, and we train the RoBERTa-base model with entire datasets.
For SciML, we evaluate \ourmethod by training or fine-tuning neural PDE solvers to learn PDEs. We use previously studied SciML models, including FNO~\citep{li2020fourier}, UNet~\citep{ronneberger2015unet} and DPOT~\citep{hao2024dpot}. We train the models on simulated solutions of PDEs: one time-independent PDE (DarcyFlow) and two time-dependent PDEs (1D and 2D CFD), with a sampling ratio from 0.6\% to 100\%.

\noindent \textbf{Baselines.} To ensure fair comparison, we use publically available pre-trained checkpoints for training, and adopt training configurations from previous works to reproduce their results. For NLP tasks, we use FT and LoRA to train the RoBERTa-base (125M) model, and we use the Adam optimizer \citep{kingma2014adam} with linear learning rate decay with warmup; for SciML tasks, we refer the experiments settings from \citep{takamoto2022pdebench}, use the Adam optimizer and schedule the base learning rate by step-wise learning rate decay. To obtain a proper hyperparameter setup, we perform grid searches on temperature parameters (learning rate, batch size). For other training configurations, we refer to existing works~\citep{liu2019roberta, hu2021lora, yang2024pde}, and find the best hyperparameters.
See Appendix~\ref{sec:subsample} and~\ref{sec:hyperparam} for details on dataset subsampling and hyperparameter configurations, respectively.

\begin{figure*}[!htb]
    \centering
    \begin{subfigure}[t]{0.56\linewidth} 
        \includegraphics[width=\linewidth]{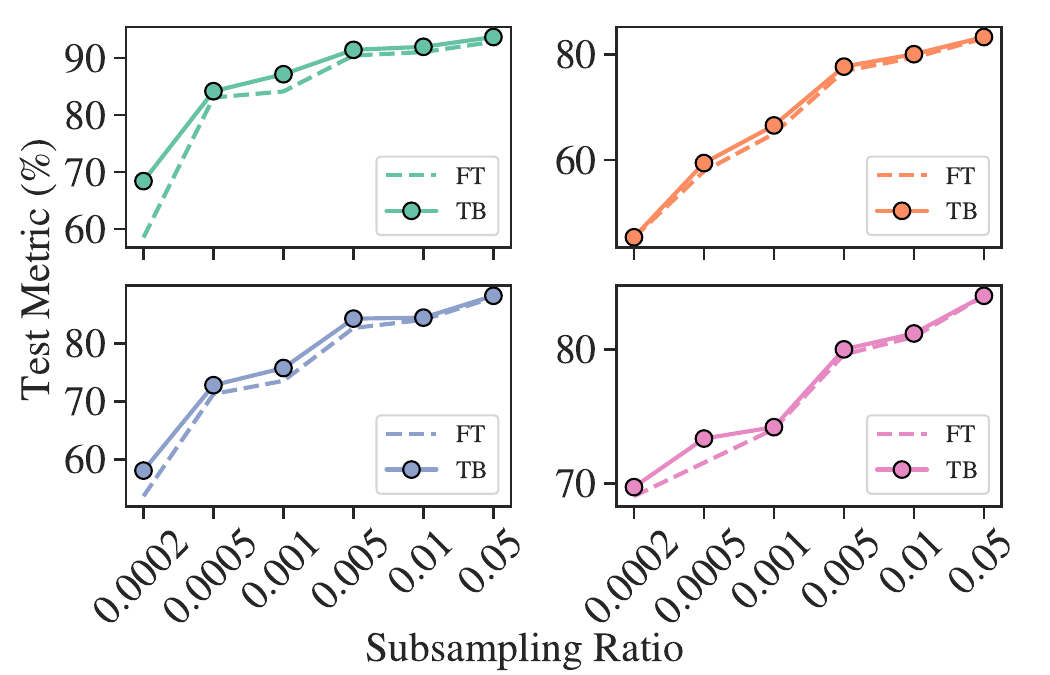}
        \caption{RoBERTa-base on GLUE datasets}
        \label{fig:nlp_improve_trend}
    \end{subfigure}
    \begin{subfigure}[t]{0.4\linewidth} 
        \includegraphics[width=\linewidth]{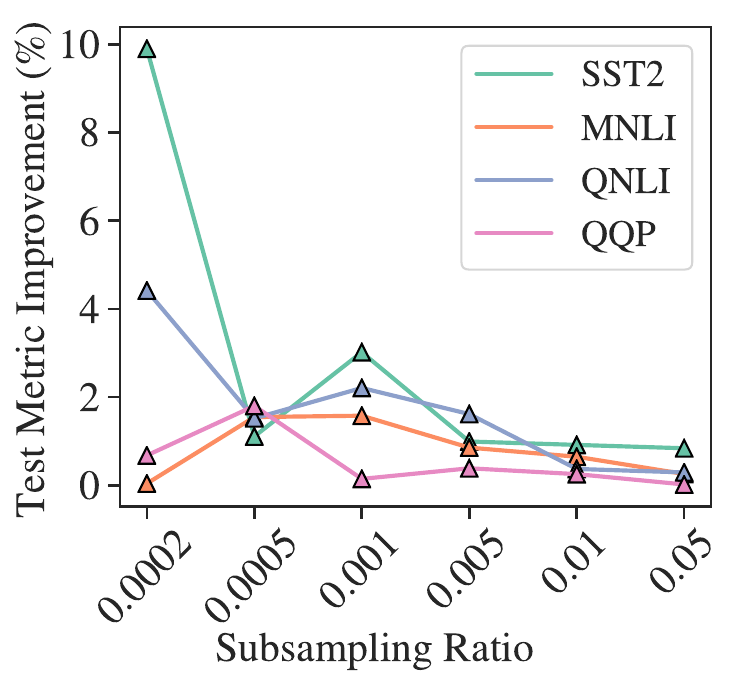}
        \caption{Trend of improvement}
        \label{fig:nlp_trend_in_one}
    \end{subfigure}
\caption{\textbf{(Main Results on LLM Fine-tuning).} \ourmethod (TB) achieves better test metric ($\uparrow$) than baseline Full Fine-tuning (FT) on GLUE tasks, especially if training data is small.~\ref{fig:nlp_improve_trend} compares test performances of baseline FT (Full Fine-tuning) and \ourmethod to train RoBERTa-base model on four larger GLUE datasets (color-coded as in~\ref{fig:nlp_trend_in_one}).~\ref{fig:nlp_trend_in_one} shows the trend of performance improvement of \ourmethod.
} 
\label{fig:improvement_trend_nlp}
\end{figure*}

\subsection{Diagnosing Layer Imbalance Using HT Metrics when Training with Limited Data }\label{sec:diagnosis}
To analyze the performance of models trained in low-data settings, we employ HT-SR theory and examine the distribution of \ALPHAHILL across different layers. Our findings reveal a strong correlation between the trend of \ALPHAHILL distribution and test performance. We use checkpoints of the RoBERTa-base model trained with subsampling ratios ranging from 0.05\% to 100\% on MNLI and QNLI dataset, and we plot the trend of test performance and block-wise STD of \ALPHAHILL, as shown in Figure~\ref{fig:trend_plot}. As test performance decreases with training data samples, we observe that the STD of \ALPHAHILL across layers increases, suggesting a more unevenly distributed \ALPHAHILL across different layers. Similar trends are also present in SciML tasks (Figure~\ref{fig:trend_plot_fno_cfd}).

Given that \ALPHAHILL is a robust predictor of model and layer quality~\citep{yang2023test, zhou2024temperature}, we propose that models trained on fewer data samples have more unevenly distributed layer qualities, this layer-wise balance becomes worse as we reduce the number of training data points. Training with more data points, on the other hand, can make the distribution of \ALPHAHILL more balanced. Therefore, when training with limited data, layer balancing is necessary for balancing the training quality of different layers.

\subsection{Improving Low-Data Training Using \ourmethod}
\label{sxn:low_data_results}

\noindent \textbf{Natural Language Understanding. } 
In Figure~\ref{fig:improvement_trend_nlp}, we report the evaluation result of fine-tuning the RoBERTa-base model with four larger GLUE datasets. We compare \ourmethod (shown as ``TB'') with Full Fine-tuning (shown as ``FT'') with different subsampling ratios. We also show the results on smaller GLUE tasks in Table~\ref{tb:glue_roberta_2}. We can see that \ourmethod consistently demonstrates performance improvement in all low-data regimes. For example, when fine-tuning on the larger SST2 dataset, \ourmethod significantly outperforms the baseline with 9.9\% improvement in test accuracy with 0.02\% subsampling ratio. Regarding the smaller RTE dataset with 50\% training data, \ourmethod can improve test accuracy by 3.13\%. The detailed results of all GLUE tasks are shown in Table~\ref{tb:glue_roberta_1} and~\ref{tb:glue_roberta_2}, in Appendix~\ref{sec:ft_glue_appendix}.

\begin{figure}
    \centering
    \includegraphics[width=0.5\linewidth]{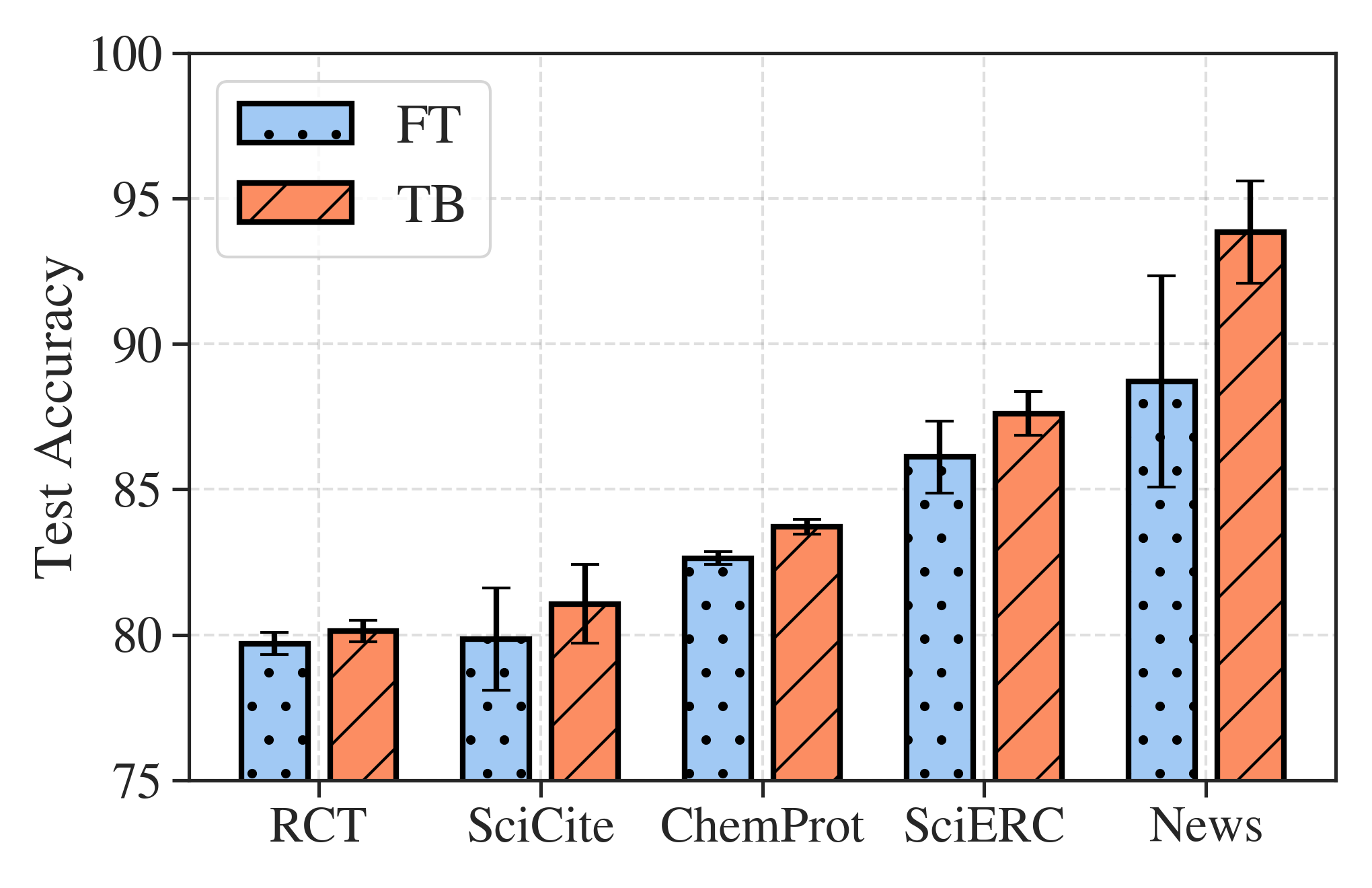}
    \caption{\textbf{Domain Specific Language Modeling.} \ourmethod demonstrates significant performance gain when training the RoBERTa-base model on five low-resource domain-specific datasets.} 
    \label{fig:domain_specific_lm}
 
\end{figure}

\noindent \textbf{Domain-specific Language Modeling. }In Figure~\ref{fig:domain_specific_lm}, we report the results of \ourmethod on five domain-specific low-resource datasets. We show that when fine-tuned on these datasets in low-data settings, \ourmethod continues to yield better test performance than the baseline method. Specifically on Hyperpartisan News dataset, \ourmethod outperforms baseline FT by 5.13\%. This indicates that \ourmethod brings significant improvement when applying to specialized language modeling domains with low resources. 

\begin{figure*}[!htb]
    \centering
    \begin{subfigure}[t]{0.56\linewidth} 
        \includegraphics[width=\linewidth]{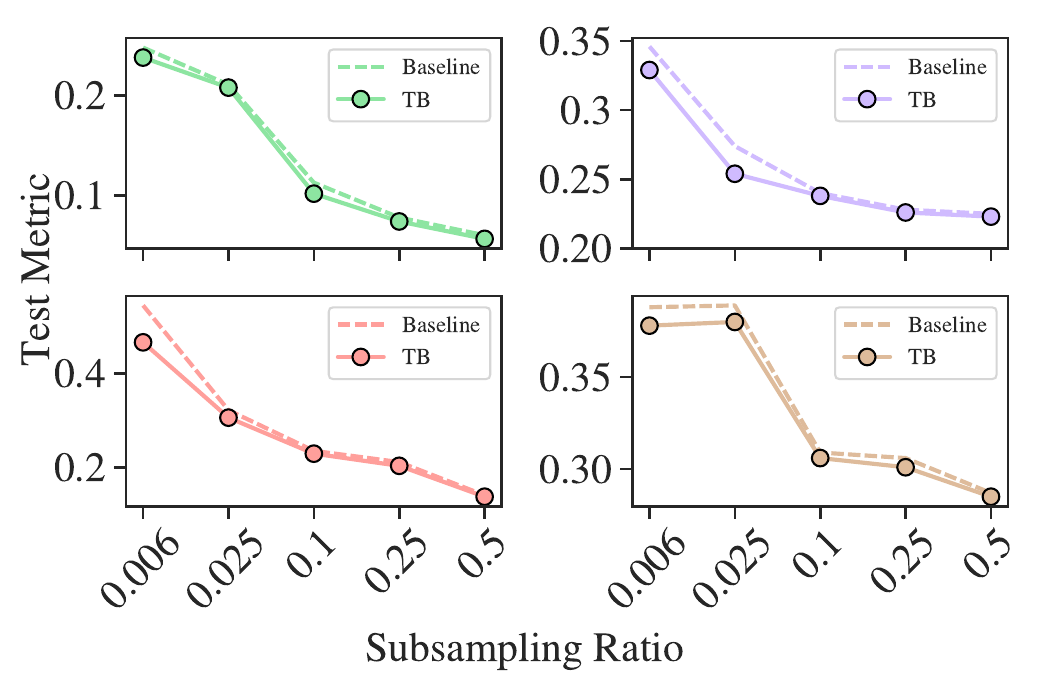}
        \caption{FNO and UNet on 1D and 2D CFD datasets}
        \label{fig:sciml_improve_trend}
    \end{subfigure}
    \begin{subfigure}[t]{0.4\linewidth} 
        \includegraphics[width=\linewidth]{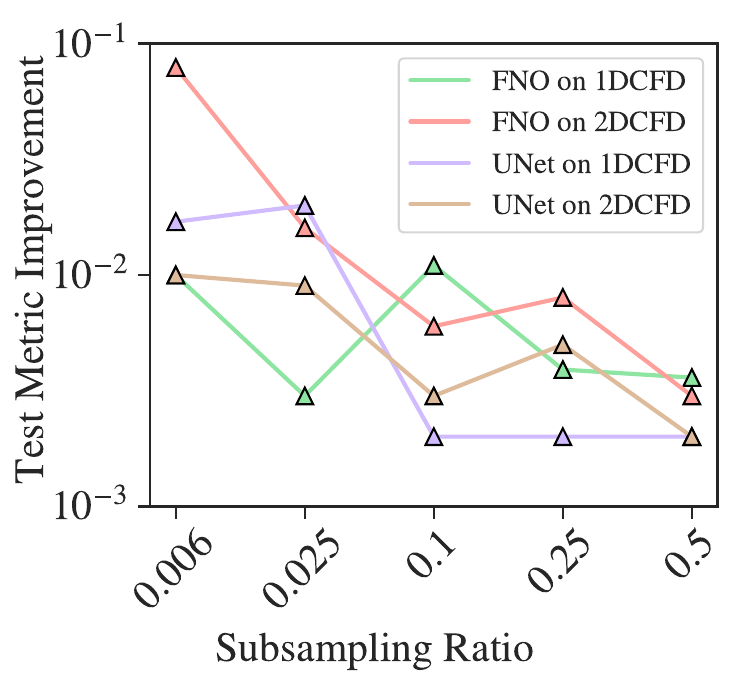}
        \caption{Trend of improvement}
        \label{fig:sciml_trend_in_one}
    \end{subfigure}
\caption{\textbf{(Main Results on PDE Learning).} \ourmethod (TB) achieves lower nRMSE($\downarrow$) than baseline method on CFD tasks, especially if subsampling ratio is small.~\ref{fig:sciml_improve_trend} compares test performances of baseline trained and \ourmethod trained FNO and UNet models on 1D and 2D CFD datasets (color-coded as in~\ref{fig:sciml_trend_in_one}).~\ref{fig:sciml_trend_in_one} demonstrates the trend of performance improvement brought by \ourmethod.}
\label{fig:improvement_trend_sciml} 
\end{figure*}

\noindent \textbf{Neural PDE Solver Training. } In Figure~\ref{fig:improvement_trend_sciml}, we report the results of training the FNO and UNet model on the 1D and 2D CFD (compressible fluid dynamics) dataset with a subsampling ratio ranging from 0.6\% to 100\%, evaluated by Normalized Root Mean Squared Error (nRMSE). The detailed results are shown in Table~\ref{tb:cfd_fno_unet}, Appendix~\ref{sec:1d_2d_cfd_appendix}. We find that \ourmethod achieves lower nRMSE compared to the baseline on all subsampling ratios. Specifically, \ourmethod reduces the nRMSE of the FNO model trained on 10.0\% of the 1DCFD dataset significantly by 9.73\% and improves the nRMSE of UNet on 2.5\% by 7.30\%. Furthermore, \ourmethod can achieve comparable performance gain to increasing the number of training data samples. For example, when solving 2DCFD problem using the UNet model with 10\% data, applying \ourmethod yields comparable performance gain to increasing the subsampling ratio to 25\%.\looseness-1

\begin{table*}[!htb]
    \centering
    \begin{minipage}{0.45\textwidth}
        \centering
        \resizebox{\textwidth}{!}{
        \begin{tabular}{c|cccc}
            \toprule
            \bf{Ratio} & 1\% & 0.5\% & 0.1\% & 0.05\% \\
            \midrule
            FT & 84.09{\scriptsize$\pm$0.36} & 82.68{\scriptsize$\pm$0.43} & 73.57{\scriptsize$\pm$0.90} & 71.31{\scriptsize$\pm$1.29} \\
            \midrule
            SAM & \textbf{85.10{\scriptsize$\pm$0.55}} & 83.35{\scriptsize$\pm$0.61} & 73.38{\scriptsize$\pm$1.48} & 71.18{\scriptsize$\pm$1.29} \\
            \midrule
            TB & 84.47{\scriptsize$\pm$0.55} & \textbf{84.30{\scriptsize$\pm$0.46}} & \textbf{75.67{\scriptsize$\pm$1.17}} & \textbf{72.65{\scriptsize$\pm$1.10}} \\
            \bottomrule
        \end{tabular}
        }
        \label{tb:sam_tb_comparison} 
    \end{minipage}
    \begin{minipage}{0.45\textwidth}
        \centering
        \resizebox{\textwidth}{!}{
        \begin{tabular}{c|cccc}
            \toprule
            \bf{Ratio} & 1\% & 0.5\% & 0.1\% & 0.05\% \\
            \midrule
            FT & 84.09{\scriptsize$\pm$0.36} & 82.68{\scriptsize$\pm$ 0.43} & 73.57{\scriptsize$\pm$0.90} & 71.31{\scriptsize$\pm$1.29} \\
            \midrule
            TB & 84.47{\scriptsize$\pm$0.55} & 83.40{\scriptsize$\pm$0.46} & 75.67{\scriptsize$\pm$1.17} & 72.65{\scriptsize$\pm$1.10} \\
            \midrule
            AdaFactor & 84.79{\scriptsize$\pm$0.37} & 83.29{\scriptsize$\pm$0.23} & 76.73{\scriptsize$\pm$0.95} & 74.09{\scriptsize$\pm$1.29} \\
            \midrule
            AdaFactor+TB & \textbf{84.81{\scriptsize$\pm$0.25}} & \textbf{84.00{\scriptsize$\pm$0.46}} & \textbf{77.75{\scriptsize$\pm$0.38}} & \textbf{76.04{\scriptsize$\pm$1.10}} \\
            \bottomrule
        \end{tabular}
        }
        \label{tb:adafactor_comparison} 
    \end{minipage}
    \caption{Comparing \ourmethod with Sharpness-Aware Minimization (SAM) and AdaFactor on RoBERTa-base model trained with QNLI dataset. For SAM, we choose hyperparameter $\rho$ in the range of \{0.5, 0.25, 0.1, 0.05\}} 
    \label{tb:optim_comparison}
\end{table*}

\noindent \textbf{Complementary Results. }To further demonstrate the generalizability of \ourmethod, we provide supplementary results on a broader range of settings in Appendix~\ref{sec:app-complement}. We first evaluate \ourmethod on more full fine-tuning and LoRA fine-tuning tasks of RoBERTa-base and LLaMA-7B, then we explore more SciML settings by training the FNO and UNet to solve CFD PDEs. We also provide statistical testing to verity the significance of our results.

\subsection{Comparison with Other Methods}\label{sxn:comparison}
Recent works have proposed optimization methods that efficiently improve low-data training especially on LLMs.
For example, Sharpness-Aware Minimization (SAM)~\citep{foret2021sharpnessaware} has been shown to effectively improve fine-tuning performance when training data is limited, by encouraging convergence to flatter local minima~\citep{bahri2022sharpnessaware}. Also, AdaFactor is a memory-efficient optimizer suitable for training large models~\citep{shazeer2018adafactor}.

We show that \ourmethod not only outperforms these methods in most low-data regimes, but can be used as an ``add-on'' method to further enhance model performance.

We compare \ourmethod with SAM and AdaFactor using RoBERTa-base model trained with QNLI on four subsampling ratios, as shown in Table~\ref{tb:optim_comparison}. We can see that when we have fewer data points, SAM achieves worse results than baseline FT. Meanwhile, \ourmethod consistently outperforms baseline FT, and achieves better results than SAM in almost all cases. For the AdaFactor optimizer, we can see that it outperforms baseline and \ourmethod in most cases. Still, when we combine \ourmethod with AdaFactor, we can achieve the best results across all low-data regimes, with at most 1.95\% test accuracy increase higher than AdaFactor alone.

\subsection{Neural PDE Fine-tuning}\label{sec:Sciml_FT}

To explore diverse scenarios in SciML, we conduct experiments on low-data fine-tuning using the 2DCFD dataset with DPOT-Tiny and DPOT-Small models. In solving PDEs, we utilize foundational models pre-trained on various fluid dynamics datasets, which are then fine-tuned on another specific physical scenario. In Table~\ref{tb:tb_nrmse_ft_sciml}, we show that \ourmethod(TB) offers better improvements compared to the baseline FT under different subsampling ratios.

The experimental settings for SciML tasks are as follows: For \ourmethod(TB) and FT, we train the models for 500 epochs with a batch size of 160 for the Tiny model and 64 for the Small model, and a dropout rate of 1e-6. We test initial learning rates among \{0.001, 0.0005, 0.00025, 0.0001, 0.00005\}. We use the Adam optimizer, and decay the learning rate by $\gamma=0.5$ every 50 epochs. The mean and standard deviation of nRMSE across 3 random seeds on the test set are reported.

\begin{table}[!htb]
    \centering
    \resizebox{0.6\columnwidth}{!}{
    \begin{tabular}{c|c|cc}
        \toprule
        \bf{Subsampling} &  &  &  \\
        \bf{Ratio} & \bf{Method} & \bf{DPOT-Tiny} & \bf{DPOT-Small}  \\
        \midrule
        5\% & FT & 1.863e-2{\scriptsize$\pm$1.067e-5} & 1.546e-2{\scriptsize$\pm$3.346e-5}  
        \\
        \rowcolor{LightBlue} & TB & \textbf{1.856e-2{\scriptsize$\pm$3.646e-5}} & \textbf{1.539e-2{\scriptsize$\pm$1.328e-5}} \\
        \midrule
        10\% & FT & 1.747e-2{\scriptsize$\pm$1.502e-5} & 1.426e-2{\scriptsize$\pm$1.157e-5}  
        \\
        \rowcolor{LightBlue} & TB & \textbf{1.730e-2{\scriptsize$\pm$1.173e-5}} & \textbf{1.415e-2{\scriptsize$\pm$1.890e-5}} \\
        \midrule
        25\% & FT & 1.543e-2{\scriptsize$\pm$4.008e-5} & 1.226e-2{\scriptsize$\pm$2.094e-5} 
        \\
        \rowcolor{LightBlue} & TB & \textbf{1.517e-2{\scriptsize$\pm$2.807e-5}} & \textbf{1.203e-2{\scriptsize$\pm$1.313e-5}} \\
        \midrule
        50\% & FT & 1.309e-2{\scriptsize$\pm$2.356e-5} & 1.025e-2{\scriptsize$\pm$2.063e-5} 
        \\
        \rowcolor{LightBlue} & TB & \textbf{1.283e-2{\scriptsize$\pm$2.494e-5}} & \textbf{1.005e-2{\scriptsize$\pm$8.860e-6}} \\
        \midrule
        100\% & FT & 1.096e-2{\scriptsize$\pm$3.875e-5} & 8.400e-3{\scriptsize$\pm$1.030e-5} 
        \\
        \rowcolor{LightBlue} & TB & \textbf{1.078e-2{\scriptsize$\pm$4.527e-5}} & \textbf{8.193e-3{\scriptsize$\pm$1.509e-5}} \\
        \bottomrule
    \end{tabular}
    }
    \caption{\texttt{TempBalance}  achieves lower nRMSE(↓) than baseline method
on SciML fine-tuning task.} \label{tb:tb_nrmse_ft_sciml} 
\end{table}

\subsection{Analysis}\label{sec:analysis}
Following section~\ref{sec:diagnosis}, we study the effectiveness of \ourmethod in overcoming low-data limitations. First, we look into the trend of improvement brought by \ourmethod, and demonstrate that layer-wise tuning like \ourmethod brings more significant improvement as we train with fewer data. Second, we investigate the distribution of \ALPHAHILL across layers, and show that \ourmethod successfully balances layer-wise training quality, resulting in a more uniform \ALPHAHILL distribution compared to the baseline method.

\noindent \textbf{Analyzing Performance Gain of \ourmethod. }As we have shown in our main results, we note that \ourmethod achieves greater performance gain as the subsampling ratio becomes lower. This trend suggests that \ourmethod is more effective as we train the model with fewer data. This trend suggests that when training data is large, model training quality is high without specific manipulations. However, if we only have a few samples, the layer-wise balancing method becomes increasingly beneficial and can significantly improve model performance.

\noindent \textbf{Analyzing \ALPHAHILL Distribution. }We compare the distribution of \ALPHAHILL between baseline FT and \ourmethod. As observed in Figure~\ref{fig:alpha_tb_std_trend}, \ourmethod consistently shows lower \ALPHAHILL variance on RoBERTa-base trained on QNLI under various subsampling ratios. Furthermore, in SciML tasks, we can see a similar trend that is more significant when we train the model from scratch (Figure~\ref{fig:alpha_std_fno_cfds}).

Following the trend shown previously in Figure~\ref{fig:trend_plot}, this finding suggests that as layer-wise training quality becomes more unevenly distributed as we train with fewer data, \ourmethod effectively balances training quality across different layers (estimated by the variance of \ALPHAHILL).

\subsection{Ablation study}
\label{sxn:ablation_study}

\noindent \textbf{Temperature Balancing with Different ESD Metrics. }Recent theoretical works have proposed several metrics that measure the shape of the ESD~\citep{martin2021implicit, martin2021predicting, yang2023test}, and we compare their performance with \ALPHAHILL in assigning layer-wise learning rates. We mainly consider two shape metrics: \SPECTRALNORM and \STABLERANK. Results are presented in Table~\ref{tb:tb_metric_comparison}. We can see that in all subsampling ratios, \ALPHAHILL continues to outperform other metrics, while other metrics may perform worse than baseline Full FT. We can conclude that \ALPHAHILL have more robust performance than other shape metrics in assigning layer-wise learning rates.

\begin{table}[!htb]
    \centering
    \resizebox{0.6\columnwidth}{!}{
    \begin{tabular}{c|cccc}
        \toprule
        \bf{Ratio} & 1\% & 0.5\% & 0.1\% & 0.05\% \\
        \midrule
        FT & 84.09{\scriptsize$\pm$0.36} & 82.68{\scriptsize$\pm$ 0.43} & 73.57{\scriptsize$\pm$0.90} & 71.31{\scriptsize$\pm$1.29} \\
        \midrule
        \SPECTRALNORM & 83.18{\scriptsize$\pm$0.41} & 81.68{\scriptsize$\pm$0.23} & 70.52{\scriptsize$\pm$5.18} & 65.79{\scriptsize$\pm$0.85} \\
        \midrule
        \STABLERANK & 83.22{\scriptsize$\pm$0.15} & 82.29{\scriptsize$\pm$0.36} & 71.87{\scriptsize$\pm$1.57} & 67.18{\scriptsize$\pm$3.71} \\
        \midrule
        \ALPHAHILL & \textbf{84.47{\scriptsize$\pm$0.55}} & \textbf{84.30{\scriptsize$\pm$0.46}} & \textbf{75.78{\scriptsize$\pm$0.47}} & \textbf{72.83{\scriptsize$\pm$1.65}} \\
        \bottomrule
    \end{tabular}
    }
    \caption{Comparing different ESD metrics used to schedule layer-wise learning rate trained with RoBERTa-base model on QNLI task. We choose \SPECTRALNORM and \STABLERANK to compare with \ALPHAHILL that we use in the \ourmethod algorithm.} \label{tb:tb_metric_comparison} 
\end{table}

\noindent \textbf{Different Learning Rate Scheduling functions. }In the \ourmethod algorithm, we choose \SIGMOID equation as our layer-wise scheduling function. To verify the superiority of \SIGMOID function, we evaluate another scheduling function \LINEARMAP, which is proven to have great performance on image classification tasks~\citep{zhou2024temperature}. The results are shown in Table~\ref{tb:tb_func_comparison}. We can see that \SIGMOID function outperforms \LINEARMAP in almost all subsampling ratios.

\begin{table}[!h]
    \centering
    \resizebox{0.6\columnwidth}{!}{
    \begin{tabular}{c|cccc}
        \toprule
        \bf{Ratio} & 1\% & 0.5\% & 0.1\% & 0.05\% \\
        \midrule
        FT & 84.09{\scriptsize$\pm$0.36} & 82.68{\scriptsize$\pm$0.43} & 73.57{\scriptsize$\pm$0.90} & 71.31{\scriptsize$\pm$1.29} \\
        \midrule
        \LINEARMAP & \textbf{84.60{\scriptsize$\pm$0.07}} & 83.87{\scriptsize$\pm$0.61} & 73.49{\scriptsize$\pm$2.92} & 72.76{\scriptsize$\pm$1.54} \\
        \midrule
        \SIGMOID & 84.47{\scriptsize$\pm$0.55} & \textbf{84.30{\scriptsize$\pm$0.46}} & \textbf{75.78{\scriptsize$\pm$0.47}} & \textbf{72.83{\scriptsize$\pm$1.65}} \\
        \bottomrule
    \end{tabular}
    }
    \caption{Comparing different Temperature Balancing scheduling algorithm on RoBERTa-base model trained with QNLI dataset.} \label{tb:tb_func_comparison} 
\end{table}

For more ablation study results on SciML tasks, please refer to Appendix~\ref{appendix:ablation_sciml}.

\section{Conclusions}
In this work, we leverage HT-SR theory to diagnose the limitations of low-data training and improve the learning rate scheduling algorithm \ourmethod to balance the training quality of different layers in low-data regimes. Our extensive experiments demonstrate that \ourmethod effectively balances layer-wise training quality and improves performance in NLP fine-tuning and SciML training. Our analysis reveals that \ourmethod achieves greater performance gain as we train with fewer data. Furthermore, the compatibility of \ourmethod makes it possible to add \ourmethod to existing optimization methods, bringing further performance improvements. We show that HT-SR theory brings useful guidance in low-data training and fine-tuning, and we expect it to be a more generalized toolbox for diagnosing model performance in more training scenarios.

\section*{Acknowledgments.}
This work is supported by DOE under Award Number DE-SC0025584, DARPA under Agreement number HR00112490441, and Dartmouth College.

\section*{Limitations}
Despite achieving improvements in NLP and SciML tasks, \ourmethod has some potential limitations.  

For computational costs, since \ourmethod dynamically reschedules learning rates during training, frequent calculations of ESD of weight matrices are required. In our work, the computation overhead of \ourmethod while training the RoBERTa-base model can take up to 25\% of the total training time: when training on 0.02\% SST2 dataset, the total training time is 265.73 seconds, in which \ourmethod takes up 65.40 seconds. This computational cost could scale up as the model size becomes larger. Since the calculation of ESD contributes to most of the computation cost (the SVD process), we will focus on improving the efficiency of measuring the Heavy-Tail structure of the ESD.  

In addition, we only discuss the scheduling of the learning rate in this work, whereas other temperature-like parameters can also influence the structure of ESD during training, such as batch size or weight decay. Therefore it would be of interest to explore how HT-SR theory can assist in acquiring a comprehensive set of hyperparameter tuning tools.

\section*{Ethics Statement}
This paper leverages HT-SR theory to design a layer-wise fine-tuning scheme for LLMs and SciML models. Our study in itself does not pose any negative societal risks or ethical concerns. On the contrary, it improves our understanding of the inner mechanisms of training NNs which can potentially aid in optimizing the amount of compute resources spent on training large NNs for wide societal use.

\clearpage

\bibliographystyle{plainnat}
\bibliography{custom}

\newpage
\appendix

\clearpage
\section*{Appendix}
\section{Potential Risks}
Our work leverages HT-SR theory as a model diagnosis tool to analyze the limitations of low-data training and fine-tuning, and help the design of an improved learning rate scheduling algorithm. We do not see any immediate negative societal impacts or ethics issues stemming from the algorithm itself. In addition, our analysis could inspire future research on diagnosing performance limitations in different scenarios, securing the safe use of LLMs.

\section{Ablation study on granularity of Learning Rate Scheduling: Per-block vs. Per-layer.}\label{appendix:schedule_ablation}
Following the discussion on scheduling method for Transformer-based models in Section~\ref{sec:tb_algorithm}, here we compare the performance of block-wise and layer-wise scheduling in RoBERTa-base model trained on QNLI dataset. Table~\ref{tb:layer_block_comp} shows that the block-wise method generally outperforms the per-layer method in different subsampling ratios. The results suggest that block-wise learning rate scheduling is a more favorable method than layer-wise scheduling when we use \ourmethod on Transformer-based models.

\begin{table}[!h]
    \centering
    \resizebox{0.8\textwidth}{!}{
    \begin{tabular}{c|ccccc}
        \toprule
        \bf{Ratio} & 5\% & 1\% & 0.5\% & 0.1\% & 0.05\% \\
        \midrule
        baseline & 87.54{\scriptsize$\pm$0.20} & 84.09{\scriptsize$\pm$0.36} & 82.68{\scriptsize$\pm$ 0.43} & 73.57{\scriptsize$\pm$0.90} & 71.31{\scriptsize$\pm$1.29} \\
        \midrule
        Layer-wise & 87.83{\scriptsize$\pm$0.23} & \textbf{84.81{\scriptsize$\pm$0.07}} & 83.78{\scriptsize$\pm$0.17} & 75.30{\scriptsize$\pm$1.72} & 70.99{\scriptsize$\pm$1.86} \\
        \midrule
        Block-wise & \textbf{88.24{\scriptsize$\pm$0.08}} & 84.47{\scriptsize$\pm$0.55} & \textbf{84.30{\scriptsize$\pm$0.46}} & \textbf{75.78{\scriptsize$\pm$0.47}} & \textbf{72.83{\scriptsize$\pm$1.65}} \\
        \bottomrule
    \end{tabular}
    }
    \caption{Comparing layer-wise and block-wise learning rate schedule trained with RoBERTa-base model on QNLI task. We choose.} \label{tb:layer_block_comp} 
\end{table}

\section{Data Subsampling}\label{sec:subsample}
To create low-data regimes, we design sets of subsampling ratios based on the size of different training datasets (see Table~\ref{tb:glue_data_num} and~\ref{tb:sciml_data_num}). For GLUE fine-tuning, we partition the datasets in GLUE into two groups: larger datasets (SST-2, MNLI, QNLI and QQP), and smaller datasets (CoLA, MRPC, STS-B and RTE). For larger datasets, we choose subsampling ratio from \{0.02\%\, 0.05\%, 0.1\%, 0.5\%, 1\%, 5\%\}, and for smaller datasets, we choose subsampling ratios from \{10\%\, 20\%, 50\%\}. For PDE solving tasks, we use the datasets from PDEBench \cite{takamoto2022pdebench} and choose different data ratios considering the training difficulty in different datasets. For DarcyFlow dataset, the range of subsampling ratio is \{0.6\%, 2.5\%, 5.0\%, 10.0\%, 100.0\%\}. For training the FNO and UNet on 1D and 2D CFD dataset, the range of subsampling ratio is \{0.6\%, 2.5\%, 10.0\%, 25.0\%, 50.0\%, 100.0\%\}. 
\begin{table}[!h]
    \centering
    \resizebox{0.8\textwidth}{!}{
    \begin{tabular}{c|cccccccc}
        \toprule
        \bf{Dataset} & \bf{SST-2} & \bf{MNLI} & \bf{QNLI} & \bf{QQP} & \bf{CoLA} & \bf{MRPC} & \bf{STS-B} & \bf{RTE} \\
        \midrule
        \bf{\# of Data} & 67K & 393K & 105K & 364K & 8.5K & 3.7K & 7K & 2.5K \\
        \bottomrule
    \end{tabular}
    }
    \caption{Number of training data samples of each GLUE tasks} \label{tb:glue_data_num}
\end{table}
\begin{table}[!h]
    \centering
    \resizebox{0.8\textwidth}{!}{
    \begin{tabular}{c|ccc}
        \toprule
        \bf{Dataset} & \bf{DarcyFlow} & \bf{1D CFD} & \bf{2D CFD }\\
        \midrule
        \bf{\# of Data} & 9K & 9K & 9K \\
        \midrule
        \bf{Parameter} & $\beta=100$ & $\eta=\zeta=0.01$, Rand periodic & $M=0.1,\eta=\zeta=0.01$, Rand periodic \\
        \bottomrule
    \end{tabular}
    }
    \caption{Number of training data samples and parameter of PDE Datasets.} \label{tb:sciml_data_num} 
\end{table}

\section{Hyperparameter Settings}\label{sec:hyperparam}
In this section, we provide detailed hyperparameter settings to reproduce the experimental results.

\subsection{Full Fine-tuning on GLUE and SuperGLUE Datasets}\label{sec:hyper-ft}
For full-finetuning, we choose to fine-tune RoBERTa-base model on GLUE and SuperGLUE datasets. For each subsampling ratio, we train using the Adam optimizer with a linear learning rate decay schedule for 10 epochs. We choose the sequence length of 128, and grid search learning rate and batch size to obtain the best results. When training on four smaller GLUE datasets (CoLA, MRPC, STSB, RTE) and SuperGLUE datasets, we search learning rate across \{1e-5, 2e-5, 3e-5\} and batch size across\{16, 32\}; when training on four larger GLUE datasets (SST2, MNLI, QNLI, QQP), the search range of learning rate and batch size are shown in Table~\ref{hyper:glue_config_lr} and~\ref{hyper:glue_config_bs} respectfully. For other hyperparameters and model configurations, we use the same settings following Liu et al.~\cite{liu2019roberta}. We report the mean over 3 random seeds for each setting, where the results for each run are taken from the best epoch.

\begin{table}[!h]
    \centering
    \resizebox{0.5\textwidth}{!}{
    \begin{tabular}{c|cccc}
        \toprule
        \bf{Dataset} & \bf{SST-2} & \bf{MNLI} & \bf{QNLI} & \bf{QQP} \\
        \midrule
        \bf{5\%} & \multicolumn{4}{c}{\{1e-5, 2e-5, 3e-5\}} \\
        \midrule
        \bf{1\%} & \multicolumn{4}{c}{\{1e-5, 2e-5, 3e-5\}} \\
        \midrule
        \bf{0.5\%} & \multicolumn{4}{c}{\{1e-5, 2e-5, 3e-5\}} \\
        \midrule
        \bf{0.1\%} & \multicolumn{4}{c}{\{1e-5, 2e-5, 3e-5\}} \\
        \midrule
        \bf{0.05\%} & \multicolumn{4}{c}{\{1e-5, 2e-5, 3e-5\}} \\
        \midrule
        \bf{0.02\%} & \{1e-5, 2e-5, 3e-5, 5e-5\} & \multicolumn{3}{|c}{\{1e-5, 2e-5, 3e-5\}} \\
        \bottomrule
    \end{tabular}
    }
    \caption{Learning rate range of training RoBERTa-base model on subsets of SST2, MNLI, QNLI and QQP datasets.} \label{hyper:glue_config_lr}
\end{table}

\begin{table}[!h]
    \centering
    \resizebox{0.5\textwidth}{!}{
    \begin{tabular}{c|cccc}
        \toprule
        \bf{Dataset} & \bf{SST-2} & \bf{MNLI} & \bf{QNLI} & \bf{QQP} \\
        \midrule
        \bf{5\%} & \multicolumn{4}{c}{\{16, 32\}} \\
        \midrule
        \bf{1\%} & \multicolumn{4}{c}{\{16, 32\}} \\
        \midrule
        \bf{0.5\%} & \multicolumn{4}{c}{\{16, 32\}} \\
        \midrule
        \bf{0.1\%} & \{4, 8, 16, 32\} & \multicolumn{3}{|c}{\{16, 32\}} \\
        \midrule
        \bf{0.05\%} & \multicolumn{3}{c|}{\{4, 8, 16, 32\}} & \{16, 32\} \\
        \midrule
        \bf{0.02\%} & \multicolumn{4}{c}{\{4, 8, 16, 32\}} \\
        \bottomrule
    \end{tabular}
    }
    \caption{Batch size range of training RoBERTa-base model on subsets of SST2, MNLI, QNLI and QQP datasets.} \label{hyper:glue_config_bs} 
\end{table}

In addition to standard training configurations, we report the hyperparameters of \ourmethod corresponding to the best results. Specifically, we report hyperparameters $s$. Note that during hyperparameter search, we find that assigning different $s$ values to layers with \ALPHAHILL higher or lower than the mean \ALPHAHILL across all layers can achieve better results, and in the tables, we show them as a pair $(s_1, s_2)$, often (2, 1).
\begin{table}[!h]
    \centering
    \resizebox{0.5\textwidth}{!}{
    \begin{tabular}{c|cccc}
        \toprule
        \bf{Dataset} & \bf{SST2} & \bf{MNLI} & \bf{QNLI} & \bf{QQP} \\
        \midrule
        \bf{5\%} & (2, 1) & 1.25 & (2, 1) & 1.25 \\
        \midrule
        \bf{1\%} & 1.25 & 1.25 & 1 & 1 \\
        \midrule
        \bf{0.5\%} & 1 & 1.25 & 1 & 1.25 \\
        \midrule
        \bf{0.1\%} & (2, 1) & 1 & 1.25 & 1.25 \\
        \midrule
        \bf{0.05\%} & 1.25 & 0.5 & 1.25 & 1 \\
        \midrule
        \bf{0.02\%} & 1.25 & 1.25 & 0.25 & 1.25 \\
        \bottomrule
    \end{tabular}
    }
    \caption{Best hyperparameter $s$ for \ourmethod of training RoBERTa-base model on subsets of SST2, MNLI, QNLI and QQP datasets.} \label{hyper:glue_config_s_1} 
\end{table}

\begin{table}[!h]
    \centering
    \resizebox{0.5\textwidth}{!}{
    \begin{tabular}{c|cccc}
        \toprule
        \bf{Dataset} & \bf{CoLA} & \bf{MRPC} & \bf{STSB} & \bf{RTE} \\
        \midrule
        \bf{50\%} & 1.25 & 1.25 & 0.75 & 1.25 \\
        \midrule
        \bf{20\%} & 1 & 1.25 & 0.5 & 0.5 \\
        \midrule
        \bf{10\%} & 1 & 1 & 1 & 1.25 \\
        \bottomrule
    \end{tabular}
    }
    \caption{Best hyperparameter $s$ for \ourmethod of training RoBERTa-base model on subsets of CoLA, MRPC, STSB and RTE datasets.} \label{hyper:glue_config_s_2} 
\end{table}

\noindent \textbf{Domain-specific Fine-tuning. }For fine-tuning on domain-specific datasets, we train the RoBERTa-base models for 10 epochs with a batch size of 16 and an initial learning rate of 3e-5. We use the AdamW optimizer and apply linear learning rate decay with a 0.06 warmup ratio. The mean and standard deviation of test accuracy across 3 random seeds on the test set are reported. 

\subsection{LoRA Fine-tuning}\label{sec:hyper-lora}
For LoRA fine-tuning, we adopt the training configurations from previous works and perform a line search around the base learning rate. For training RoBERTa-base model on GLUE datasets, we follow Hu et al~\cite{hu2021lora}. and evaluate learning rate at 2e-4 and 6e-4 around the base learning rate (4e-4 or 5e-5). For LLaMA-7B on ScienceQA, we trained with AdamW optimizer for 50 epochs, and search the best learning rate in the range of \{2e-4, 3e-4, 4e-4\}. We set the cutoff length as 256 and batch size as 128. For LoRA adapters, we set the rank to 8, LoRA alpha to 16, and LoRA dropout to 0.05.

\subsection{Neural PDE Solving}\label{sec:hyper-sciml} For SciML, we referred to PDEBench\cite{takamoto2022pdebench} for the hyperparameter settings and selected the appropriate learning rate, weight decay and batch size using a grid search method to make baseline models achieve good performances. For each subsampling ratio, we train the models with the Adam optimizer, scheduling the base learning rate by decaying the learning rate by $\gamma=0.5$ every 100 epochs. We chose to train the models for enough epochs to ensure that the trained models were close to a convergent state. For the hyperparameter $s$ in \ourmethod, we choose from the range $\{0.125, 0.25, 0.5, 0.75, 1.0, 1.25, 1.5\}$. 

For training the FNO and UNet on DarcyFlow ($\beta=100$), the search range of leanring rate and the selected weight decay are displayed in Table~\ref{hyper:darcy_fno} and the batch size is 50. 

\begin{table}[!htb]
    \centering
    \resizebox{0.8\textwidth}{!}{
    \begin{tabular}{c|cccc}
        \toprule
        \bf{Model} & \multicolumn{2}{c}{\bf{FNO}} & \multicolumn{2}{c}{\bf{UNet}} \\
        \midrule
        \bf{Hyperparameters} & \bf{Learning Rate} & \bf{Weight Decay} & \bf{Learning Rate} & \bf{Weight Decay} \\
        \midrule
        \bf{100\%} & {\{5e-3, 1e-2, 1.5e-2\}} & 1e-6
        & \{2.5e-4, 5e-4, 1e-3\} & 1e-7 \\
        
        \midrule
        \bf{10.0\%} & \{1.5e-2, 2.5e-2, 5e-2\} & 1e-4 
        & \{5e-3, 1e-2, 2.5e-2\} & 1e-4 \\
        
        \midrule
        \bf{5.0\%} &  \{1.5e-2, 2.5e-2, 5e-2\}   & 1e-3 
        & \{5e-3, 1e-2, 2.5e-2\} & 1e-3 \\
        
        \midrule
        \bf{2.5\%} & \{1.5e-2, 2.5e-2, 5e-2\} & 1e-3 
        & \{1.5e-2, 2.5e-2, 5e-2\} & 1e-3 \\
        
        \midrule
        \bf{0.6\%} &  \{1.5e-2, 2.5e-2, 5e-2\}  & 1e-2 
        & \{2.5e-2, 5e-2, 1e-1\} & 1e-3 \\
        \bottomrule
    \end{tabular}
    }
    \caption{Learning rate range and the selected weight decay of training FNO and UNet model on subsets of DarcyFlow($\beta=100.0$) dataset.} \label{hyper:darcy_fno}
    
\end{table}

When training the FNO on 1D and 2D CFD datasets, the search range of learning rate and the selected weight decay is shown in Table~\ref{hyper:cfd_fno}. The batch size for the subsampling ratio \{100\%, 50.0\%, 25.0\%, 10.0\%\} in training on 1DCFD is 25 and 10 for \{2.5\%, 0.6\%\}, while on the 2DCFD dataset the batch size is 20. 

\begin{table}[!htb]
    \centering
    \resizebox{0.8\textwidth}{!}{
    \begin{tabular}{c|cccc}
        \toprule
        \bf{Dataset} & \multicolumn{2}{c}{\bf{1DCFD}} & \multicolumn{2}{c}{\bf{2DCFD}} \\
        \midrule
        \bf{Hyperparameters} & \bf{Learning Rate} & \bf{Weight Decay} & \bf{Learning Rate} & \bf{Weight Decay} \\
        \midrule
        \bf{100\%} & {\{2.5e-3, 5e-3, 1e-2\}} & 1e-2
        & \{1e-3, 2.5e-3, 5e-3\} & 1e-4 \\
        
        \midrule
        \bf{50.0\%} & {\{2.5e-3, 5e-3, 1e-2\}} & 1e-2 
        & \{1e-3, 2.5e-3, 5e-3\} & 1e-4 \\

        \midrule
        \bf{25.0\%} &  \{2.5e-3, 5e-3, 1e-2\}   & 1e-2 
        & \{1e-3, 2.5e-3, 5e-3\} & 1e-4 \\
        
        \midrule
        \bf{10.0\%} &  \{2.5e-3, 5e-3, 1e-2\}   & 1e-2 
        & \{1e-3, 2.5e-3, 5e-3\} & 1e-4 \\
        
        \midrule
        \bf{2.5\%} & \{2.5e-3, 5e-3, 1e-2\} & 1e-1 
        & \{1e-3, 2.5e-3, 5e-3\} & 1e-4 \\
        
        \midrule
        \bf{0.6\%} &  \{1e-3, 2.5e-3, 5e-3\}  & 1e-1 
        & \{2.5e-3, 5e-3, 1e-2\} & 1e-4 \\
        \bottomrule
    \end{tabular}
    }
    \caption{Learning rate range and the selected weight decay of training FNO model on subsets of 1D and 2D CFD datasets.} \label{hyper:cfd_fno}
\end{table}

Table~\ref{hyper:cfd_unet} demonstrates the properly chosen weight decay and the learning rate range of training UNet on 1D and 2D CFD datasets. The batch size for the subsampling ratio \{100\%, 50.0\%, 25.0\%\} in training on 1DCFD is 100, for \{10.0\%, 2.5\%\} is 50, and for\{0.6\%\} is 25, while on the 2DCFD dataset the batch size is 20. 

\begin{table}[!htb]
    \centering
    \resizebox{0.8\textwidth}{!}{
    \begin{tabular}{c|cccc}
        \toprule
        \bf{Dataset} & \multicolumn{2}{c}{\bf{1DCFD}} & \multicolumn{2}{c}{\bf{2DCFD}} \\
        \midrule
        \bf{Hyperparameters} & \bf{Learning Rate} & \bf{Weight Decay} & \bf{Learning Rate} & \bf{Weight Decay} \\
        \midrule
        \bf{100\%} & {\{5e-3, 1e-2, 2.5e-2\}} & 1e-5
        & \{1e-2, 2.5e-2, 5e-2\} & 1e-3 \\
        
        \midrule
        \bf{50.0\%} & {\{5e-3, 1e-2, 2.5e-2\}} & 1e-1 
        & \{2.5e-3, 5e-3, 1e-2\} & 1e-1 \\

        \midrule
        \bf{25.0\%} & {\{5e-3, 1e-2, 2.5e-2\}} & 1e-1 
        & \{2.5e-3, 5e-3, 1e-2\} & 1e-1 \\
        
        \midrule
        \bf{10.0\%} &  \{5e-3, 1e-2, 2.5e-2\}   & 1e-1 
        & \{2.5e-3, 5e-3, 1e-2\} & 1e-1 \\
        
        \midrule
        \bf{2.5\%} & \{2.5e-2, 5e-2, 1e-1\} & 1e-1 
        & \{5e-3, 1e-2, 2.5e-2\} & 1e-1 \\
        
        \midrule
        \bf{0.6\%} &  \{2.5e-2, 5e-2, 1e-1\}  & 1e-1 
        & \{2.5e-2, 5e-2, 1e-1\} & 1e-1 \\
        \bottomrule
    \end{tabular}
    }
    \caption{Learning rate range and the selected weight decay of training UNet model on subsets of 1D and 2D CFD datasets.} \label{hyper:cfd_unet}
\end{table}

\section{Complementary Results}\label{sec:app-complement}
In this section, we first provide detailed results discussed in Section~\ref{sxn:low_data_results} in the paper, then further evaluate \ourmethod on NLP and SciML training tasks. Also in Section~\ref{sec:statistical_testing}, we provide statistical testing results to demonstrate the significance of improvement brought by \ourmethod. First, in~\ref{sec:ft_glue_appendix} and~\ref{sec:1d_2d_cfd_appendix} we show detailed results of GLUE full fine-tuning and two time-dependent PDEs discussed in Section~\ref{sxn:low_data_results}. Second, we present complementary results of \ourmethod on fine-tuning RoBERTa-base model on SuperGLUE and SQuAD datasets in~\ref{sec:superglue_appendix}. Then, we apply \ourmethod to LoRA fine-tuning, and show the results of LoRA fine-tuning of RoBERTa-base model on GLUE tasks in~\ref{sec:lora_glue_appendix}, and LLaMA-7B model on ScienceQA in~\ref{sec:llama_qa_appendix}. Afterwards, we evaluate \ourmethod on solving DarcyFlow PDEs with FNO and UNet model in~\ref{sec:darcy_appendix}.

\subsection{Detailed Fine-tuning Results on GLUE Datasets}~\label{sec:ft_glue_appendix}
In Table~\ref{tb:glue_roberta_1} and~\ref{tb:glue_roberta_2}, we show the full results of fine-tuning RoBERTa-base model on GLUE datasets, corresponding to Figure~\ref{fig:improvement_trend_nlp} and the discussions in Section~\ref{sxn:low_data_results}.

\subsection{Statistical Testing on the Significance of Improvement}\label{sec:statistical_testing}
We perform statistical testing to verify the effectiveness of our algorithm compared to baseline methods. We define the Null Hypothesis (H0) as ``There is no significant difference in performance between our algorithm and the baseline'', and the Alternative Hypothesis (H1) as ``Our algorithm performs significantly better than the baseline''. We run experiments of training RoBERTa-base on SST-2 with different subsampling ratios for 10 random seeds and perform t-tests on the results. We present the results in the table below:

\begin{table}[!h]
    \centering
    \resizebox{0.5\textwidth}{!}{
    \begin{tabular}{c|ccccc}
        \toprule
        \bf{Ratio} & 0.02\% & 0.1\% & 0.5\% & 1\% & 5\% \\
        \midrule
        \bf{P-value} & 3.85$e^{-9}$ & 0.13 & 0.003 & 0.003 & 4.06$e^{-5}$ \\
        \bottomrule
    \end{tabular}
    }
    \caption{Statistical testing results on RoBERTa-base model trained with different subsampling ratios of the SST-2 dataset.} \label{tb:testing}
\end{table}

\begin{table}[!h]
    \centering
    \resizebox{0.8\textwidth}{!}{
    \begin{tabular}{c|ccccc}
        \toprule
        \bf{Subsampling} & & & & & \\
        \bf{Ratio} & 1\% & 5\% & 10\% & 20\% & 50\% \\
        \midrule
        FT & 45.84{\scriptsize$\pm$2.26} & 79.49{\scriptsize$\pm$0.22} & 86.88{\scriptsize$\pm$0.12} & 88.56{\scriptsize$\pm$0.14} & 90.97{\scriptsize$\pm$0.15} \\
        \midrule
        \rowcolor{LightBlue}TB & \textbf{48.91{\scriptsize$\pm$1.27}} & \textbf{81.18{\scriptsize$\pm$0.07}} & \textbf{88.08{\scriptsize$\pm$0.05}} & \textbf{89.49{\scriptsize$\pm$0.20}} &
        \textbf{91.16{\scriptsize$\pm$0.03}} 
        \\
        \bottomrule
    \end{tabular}
    }
    \caption{Test accuracy (\%) on SQuAD v1.1 dataset of ROBERTa-base model trained with different subsampled training sets.} \label{tb:squad}
\end{table}

\begin{table*}[!thb]
    \centering
    \resizebox{0.8\textwidth}{!}{
    \begin{tabular}{c|c|ccccc}
        \toprule
        \bf{Subsampling} &  &  &  &  &  & \\
        \bf{Ratio} & \bf{Method} & \bf{SST-2} & \bf{MNLI} & \bf{QNLI} & \bf{QQP} & \bf{Avg.} \\
        \midrule
        0.02\% & FT & 58.49{\scriptsize$\pm$10.96} & 45.28{\scriptsize$\pm$0.62} & 53.69{\scriptsize$\pm$0.44} & 69.04{\scriptsize$\pm$0.19} & 56.63 \\
        \rowcolor{LightBlue} & TB & \textbf{68.39{\scriptsize$\pm$3.21}} & \textbf{45.32{\scriptsize$\pm$1.31}} & \textbf{58.11{\scriptsize$\pm$6.29}} & \textbf{69.72{\scriptsize$\pm$0.70}} & \textbf{60.39($\uparrow$3.76)} \\
        \midrule
        0.05\% & FT & 83.07{\scriptsize$\pm$0.66} & 57.87{\scriptsize$\pm$1.14} & 71.31{\scriptsize$\pm$1.29} & 71.55{\scriptsize$\pm$1.25} & 70.95 \\
        \rowcolor{LightBlue} & TB & \textbf{84.17{\scriptsize$\pm$0.25}} & \textbf{59.42{\scriptsize$\pm$1.90}} & \textbf{72.83{\scriptsize$\pm$1.65}} & \textbf{73.35{\scriptsize$\pm$1.43}} & \textbf{72.44($\uparrow$1.49)} \\
        \midrule
        0.1\% & FT & 84.13{\scriptsize$\pm$1.97} & 64.99{\scriptsize$\pm$2.39} & 73.57{\scriptsize$\pm$0.90} & 74.05{\scriptsize$\pm$0.94} & 74.19 \\
        \rowcolor{LightBlue} & TB & \textbf{87.16{\scriptsize$\pm$0.81}} & \textbf{66.57{\scriptsize$\pm$2.51}} & \textbf{75.78{\scriptsize$\pm$0.47}} & \textbf{74.20{\scriptsize$\pm$0.61}} & \textbf{75.93($\uparrow$1.74)} \\
        \midrule
        0.5\% & FT & 90.44{\scriptsize$\pm$0.46} & 76.88{\scriptsize$\pm$0.33} & 82.68{\scriptsize$\pm$0.43} & 79.61{\scriptsize$\pm$0.24} & 82.40 \\
        \rowcolor{LightBlue} & TB & \textbf{91.44{\scriptsize$\pm$0.42}} & \textbf{77.73{\scriptsize$\pm$0.47}} & \textbf{84.30{\scriptsize$\pm$0.46}} & \textbf{80.00{\scriptsize$\pm$0.21}} & \textbf{83.37($\uparrow$0.97)} \\
        \midrule
        1\% & FT & 91.06{\scriptsize$\pm$0.16} & 79.45{\scriptsize$\pm$0.22} & 84.09{\scriptsize$\pm$0.36} & 80.93{\scriptsize$\pm$0.31} & 83.88 \\
        \rowcolor{LightBlue} & TB & \textbf{91.97{\scriptsize$\pm$0.48}} & \textbf{80.10{\scriptsize$\pm$0.25}} & \textbf{84.47{\scriptsize$\pm$0.55}} & \textbf{81.18{\scriptsize$\pm$0.22}} & \textbf{84.43($\uparrow$0.55)} \\
        \midrule
        5\% & FT & 92.85{\scriptsize$\pm$0.24} & 83.10{\scriptsize$\pm$0.02} & 87.94{\scriptsize$\pm$0.08} & 83.98{\scriptsize$\pm$0.04} & 86.97 \\
        \rowcolor{LightBlue} & TB & \textbf{93.69{\scriptsize$\pm$0.16}} & \textbf{83.36{\scriptsize$\pm$0.15}} & \textbf{88.24{\scriptsize$\pm$0.08}} & \textbf{84.00{\scriptsize$\pm$0.15}} & \textbf{87.32($\uparrow$0.35)} \\
        \bottomrule
    \end{tabular}
    }
    \caption{Evaluation results of RoBERTa-base model trained on larger GLUE tasks. We compare \ourmethod (TB) with Full Fine-tuning (FT) trained with Adam optimizer and linear learning rate decay. The tasks and their corresponding evaluation metrics are: SST-2 (accuracy, $\uparrow$), MNLI (accuracy, $\uparrow$), QNLI (accuracy, $\uparrow$) and QQP (combined score of F1 score and accuracy, $\uparrow$)} \label{tb:glue_roberta_1}
\end{table*}

\begin{table*}[!ht]
    \centering
    \resizebox{0.8\textwidth}{!}{
    \begin{tabular}{c|c|ccccc}
        \toprule
        \bf{Subsampling} &  &  &  &  &  & \\
        \bf{Ratio} & \bf{Method} & \bf{CoLA} & \bf{MRPC} & \bf{STSB} & \bf{RTE} & \bf{Avg.} \\
        \midrule
        10\% & FT & 49.01{\scriptsize$\pm$1.63} & 81.29{\scriptsize$\pm$1.61} & 84.36{\scriptsize$\pm$1.03} & 59.69{\scriptsize$\pm$0.45} & 68.59 \\
        \rowcolor{LightBlue} & TB & \textbf{50.34{\scriptsize$\pm$0.91}} & \textbf{81.70{\scriptsize$\pm$1.61}} & \textbf{86.04{\scriptsize$\pm$0.80}} & \textbf{60.53{\scriptsize$\pm$1.78}} & \textbf{69.65($\uparrow$1.06)} \\
        \midrule
        20\% & FT & 49.50{\scriptsize$\pm$2.08} & 84.64{\scriptsize$\pm$0.50} & 87.45{\scriptsize$\pm$0.25} & 66.07{\scriptsize$\pm$0.88} & 71.92 \\
        \rowcolor{LightBlue} & TB & \textbf{51.28{\scriptsize$\pm$0.73}} & \textbf{85.86{\scriptsize$\pm$0.61}} & \textbf{88.39{\scriptsize$\pm$0.55}} & \textbf{67.27{\scriptsize$\pm$0.34}} & \textbf{73.13($\uparrow$1.21)} \\
        \midrule
        50\% & FT & 56.78{\scriptsize$\pm$1.96} & 87.66{\scriptsize$\pm$0.42} & 90.12{\scriptsize$\pm$0.20} & 71.48{\scriptsize$\pm$1.35} & 76.51 \\
        \rowcolor{LightBlue} & TB & \textbf{58.60{\scriptsize$\pm$0.74}} & \textbf{88.40{\scriptsize$\pm$0.42}} & \textbf{90.24{\scriptsize$\pm$0.06}} & \textbf{74.85{\scriptsize$\pm$1.78}} & \textbf{78.02($\uparrow$1.51)} \\
        \bottomrule
    \end{tabular}
    }
    \caption{Evaluation results of RoBERTa-base trained on smaller GLUE tasks using full fine-tuning. We compare \ourmethod with baseline FT (Full Fine-tuning) on: CoLA (Matthews Correlation, $\uparrow$), MRPC (combined score of F1 score and accuracy, $\uparrow$), STS-B (combined score of Pearson and Spearman Rank, $\uparrow$), and RTE (Accuracy, $\uparrow$) \looseness-1} \label{tb:glue_roberta_2}
\end{table*}

\subsection{Full Fine-tuning on SuperGLUE and SQuAD}\label{sec:superglue_appendix}
\noindent \textbf{SuperGLUE. }In Table~\ref{tb:superglue_roberta}, we present the results of applying \ourmethod on training RoBERTa-base model on SuperGLUE tasks. The tasks
and their corresponding evaluation metrics are: BoolQ (Accuracy), RTE (Accuracy), CB (Accuracy and F1), WiC (Accuracy), MultiRC (F1 and Exact Match (EM)), COPA (Accuracy). We can see that \ourmethod effectively increases test performance in most cases, and archives significant overall improvement. Specifically, \ourmethod achieves 7.14\% performance gain when training on 50\% CB dataset. \ourmethod can also improve the overall mean performance by 1.65\% when trained with 50\% data.

\noindent \textbf{SQuAD. }In Table~\ref{tb:squad}, we present the results of applying \ourmethod on training RoBERTa-base model on SQuAD (v1.1) dataset across five subsampling ratios: {1\%, 5\%, 10\%, 20\%, 50\%}. We train the model for 10 epochs with learning rate 2e-5 and a batch size of 24 using the AdamW optimizer with a warmup rate of 0.06 and linear learning rate decay. We follow the detailed hyperparameter settings from~\citep{liu2019roberta}. The mean and standard deviation of test accuracy across 3 random seeds on the test set are reported. We observe that \ourmethod continues to achieve better test performance than baseline FT, and significantly outperforms baseline FT in low-data regimes: when trained on 1\% data of SQuAD, \ourmethod increases the test accuracy by 3.07\%.

\subsection{Detailed Results on 1D and 2D CFD Datasets}\label{sec:1d_2d_cfd_appendix}

\begin{table*}
    \centering
    \resizebox{0.8\textwidth}{!}{
    \begin{tabular}{c|c|cccc}
        \toprule
        \bf{Subsampling} & \bf{Model} & \multicolumn{2}{c}{\bf{FNO}} & \multicolumn{2}{c}{\bf{UNet}}  \\
        \cmidrule(l{0pt}r{0pt}){2-6}
        \rowcolor{LightBlue} \bf{Ratio} & \bf{Dataset} & \bf{1DCFD} & \bf{2DCFD} & \bf{1DCFD} & \bf{2DCFD} \\
        \midrule
        & Baseline & 5.02e-02{\scriptsize$\pm$4.43e-03} & 1.23e-01{\scriptsize$\pm$7.44e-03} & 2.08e-01{\scriptsize$\pm$1.71e-02} & 2.96e-01{\scriptsize$\pm$7.05e-03} \\
        \rowcolor{LightBlue} 100\% & TB & \textbf{4.74e-02{\scriptsize$\pm$6.57e-04}} & \textbf{1.16e-01{\scriptsize$\pm$4.29e-03}} & \textbf{1.91e-01{\scriptsize$\pm$1.59e-02}} & \textbf{2.90e-01{\scriptsize$\pm$1.94e-03}} \\
        & Error Reduced & 5.58\% & 5.69\% & 8.17\% & 2.03\% \\
        \midrule
        
        & Baseline & 6.04e-02{\scriptsize$\pm$3.17e-03}  &   1.40e-01{\scriptsize$\pm$4.68e-03}  & 2.25e-01{\scriptsize$\pm$2.24e-03} & 2.87e-01{\scriptsize$\pm$6.49e-03} \\
        \rowcolor{LightBlue} 50.0\% & TB & \textbf{5.68e-02{\scriptsize$\pm$2.28e-03}}  & \textbf{1.37e-01{\scriptsize$\pm$3.53e-03}} & \textbf{2.23e-01{\scriptsize$\pm$1.24e-03}} & \textbf{2.85e-01{\scriptsize$\pm$5.64e-04}} \\
        & Error Reduced & 5.96\% & 2.14\% & 0.89\% & 0.70\% \\
        \midrule
        
        & Baseline & 7.81e-02{\scriptsize$\pm$3.79e-03} &  2.11e-01{\scriptsize$\pm$3.27e-03} & 2.28e-01{\scriptsize$\pm$1.79e-03} &  3.06e-01{\scriptsize$\pm$1.77e-03}\\
        \rowcolor{LightBlue} 25.0\% & TB & \textbf{7.42e-02{\scriptsize$\pm$1.87e-03}} & \textbf{2.03e-01{\scriptsize$\pm$5.54e-03}} & \textbf{2.26e-01{\scriptsize$\pm$1.52e-03}} & \textbf{3.01e-01{\scriptsize$\pm$1.63e-03}} \\
        & Error Reduced & 4.99\% & 3.79\% & 0.88\% & 1.97\% \\
        \midrule
        
        & Baseline & 1.13e-01{\scriptsize$\pm$4.79e-03} & 2.35e-01{\scriptsize$\pm$1.61e-03} & 2.40e-01{\scriptsize$\pm$2.42e-03} & 3.09e-01{\scriptsize$\pm$1.92e-03}\\
        \rowcolor{LightBlue} 10.0\% & TB & \textbf{1.02e-01{\scriptsize$\pm$1.88e-03}} & \textbf{2.29e-01{\scriptsize$\pm$1.41e-03}} & \textbf{2.38e-01{\scriptsize$\pm$2.00e-04}} & \textbf{3.06e-01{\scriptsize$\pm$2.96e-03}}\\
        & Error Reduced & 9.73\% & 2.55\% & 0.83\% & 0.97\% \\
        \midrule
        
        & Baseline & 2.11e-01{\scriptsize$\pm$2.79e-03}  &  3.22e-01{\scriptsize$\pm$5.37e-03} & 2.74e-01{\scriptsize$\pm$2.88e-02} & 3.89e-01{\scriptsize$\pm$3.77e-02} \\
        \rowcolor{LightBlue} 2.5\% & TB & \textbf{2.08e-01{\scriptsize$\pm$5.25e-03}} &  \textbf{3.06e-01{\scriptsize$\pm$1.15e-02}} & \textbf{2.54e-01{\scriptsize$\pm$4.61e-03}} & \textbf{3.80e-01{\scriptsize$\pm$1.76e-02}} \\
        & Error Reduced & 1.42\% & 4.97\% & 7.30\% & 2.31\% \\
        \midrule

        & Baseline & 2.48e-01{\scriptsize$\pm$3.35e-03} &  5.46e-01{\scriptsize$\pm$2.20e-02} & 3.46e-01{\scriptsize$\pm$4.15e-03} & 3.88e-01{\scriptsize$\pm$2.15e-02} \\
        \rowcolor{LightBlue} 0.6\% & TB & \textbf{2.38e-01{\scriptsize$\pm$2.84e-03}} &  \textbf{4.67e-01{\scriptsize$\pm$2.85e-02}} & \textbf{3.29e-01{\scriptsize$\pm$1.87e-02}} & \textbf{3.78e-01{\scriptsize$\pm$2.78e-02}} \\
        & Error Reduced & 4.03\% & 14.47\% & 4.91\% & 2.58\% \\
        \bottomrule
    \end{tabular}
    }
    \caption{Evaluation results of FNO and UNet model trained on 1D and 2D CFD datasets. We compare our method (TB) with the baseline. The evaluation metric is nRMSE ($\downarrow$).} \label{tb:cfd_fno_unet} 
\end{table*}

In Table~\ref{tb:cfd_fno_unet}, we present the detailed results of training FNO and UNet model on 1D and 2D CFD datasets, corresponding to Figure~\ref{fig:improvement_trend_sciml} and the discussions in Section~\ref{sxn:low_data_results}.

\begin{table*}[!ht]
    \centering
    \resizebox{0.8\textwidth}{!}{
    \begin{tabular}{c|c|ccccccc}
        \toprule
        \bf{Subsampling} &  &  &  &  &  &  & \\
        \bf{Ratio} & \bf{Method} & \bf{BoolQ} & \bf{RTE} & \bf{CB} & \bf{WiC} & \bf{MultiRC} & \bf{COPA} & \bf{Avg.} \\
        \midrule
        10\% & FT & 64.97{\scriptsize$\pm$2.58} & 62.57{\scriptsize$\pm$1.68} & 68.45{\scriptsize$\pm$2.23} & 62.80{\scriptsize$\pm$3.00} & 32.95{\scriptsize$\pm$0.33} & 54.67{\scriptsize$\pm$0.47} & 57.73 \\
        \rowcolor{LightBlue} & TB & \textbf{65.95{\scriptsize$\pm$2.17}} & \textbf{62.69{\scriptsize$\pm$1.19}} & \textbf{69.64{\scriptsize$\pm$1.46}} & \textbf{63.43{\scriptsize$\pm$1.90}} & \textbf{33.22{\scriptsize$\pm$0.47}} & \textbf{58.33{\scriptsize$\pm$2.62}} & \textbf{58.88($\uparrow$1.15)} \\
        \midrule
        20\% & FT & 69.93{\scriptsize$\pm$2.01} & 67.87{\scriptsize$\pm$1.64} & 72.61{\scriptsize$\pm$0.84} & \textbf{67.14{\scriptsize$\pm$0.98}} & 34.92{\scriptsize$\pm$0.88} & 57.00{\scriptsize$\pm$2.16} & 61.58 \\
        \rowcolor{LightBlue} & TB & \textbf{71.80{\scriptsize$\pm$1.92}} & \textbf{70.04{\scriptsize$\pm$1.35}} & 72.61{\scriptsize$\pm$0.84} & 66.67{\scriptsize$\pm$1.74} & \textbf{35.00{\scriptsize$\pm$0.16}} & \textbf{59.33{\scriptsize$\pm$6.13}} & \textbf{62.58($\uparrow$1.00)} \\
        \midrule
        50\% & FT & 76.73{\scriptsize$\pm$0.49} & 74.84{\scriptsize$\pm$0.90} & 77.38{\scriptsize$\pm$2.23} & 68.44{\scriptsize$\pm$2.50} & 35.77{\scriptsize$\pm$0.92} & 58.67{\scriptsize$\pm$1.25} & 65.29 \\
        \rowcolor{LightBlue} & TB & \textbf{76.85{\scriptsize$\pm$0.13}} & 74.84{\scriptsize$\pm$1.62} & \textbf{84.52{\scriptsize$\pm$0.03}} & \textbf{70.32{\scriptsize$\pm$1.10}} & \textbf{36.44{\scriptsize$\pm$0.59}} & 58.67{\scriptsize$\pm$2.87} & \textbf{66.94($\uparrow$1.65)} \\
        \bottomrule
    \end{tabular}
    }
    \caption{Evaluation results of RoBERTa-base model trained on SuperGLUE tasks using full fine-tuning.} \label{tb:superglue_roberta}
\end{table*}

\begin{table*}
    \centering
    \resizebox{0.8\textwidth}{!}{
    \begin{tabular}{c|c|ccccc}
        \toprule
        \bf{Subsampling} &  &  &  &  &  & \\
        \bf{Ratio} & \bf{Method} & \bf{SST-2} & \bf{MNLI} & \bf{QNLI} & \bf{QQP} & \bf{Avg.} \\
        \midrule
        0.02\% & LoRA & 66.82{\scriptsize$\pm$0.81} & 37.93{\scriptsize$\pm$0.89} & 51.58{\scriptsize$\pm$0.29} & 61.18{\scriptsize$\pm$2.72} & 54.38 \\
        \rowcolor{LightBlue} & LoRA+TB & \textbf{70.11{\scriptsize$\pm$0.84}} & \textbf{39.39{\scriptsize$\pm$1.84}} & \textbf{51.93{\scriptsize$\pm$0.41}} & \textbf{63.77{\scriptsize$\pm$0.99}} & \textbf{56.3($\uparrow$1.92)} \\
        \midrule
        0.05\% & LoRA & \textbf{82.03{\scriptsize$\pm$1.33}} & 54.74{\scriptsize$\pm$0.57} & 54.91{\scriptsize$\pm$0.41} & 67.80{\scriptsize$\pm$0.62} & 64.87 \\
        \rowcolor{LightBlue} & LoRA+TB & 81.77{\scriptsize$\pm$1.97} & \textbf{55.19{\scriptsize$\pm$0.97}} & \textbf{59.93{\scriptsize$\pm$1.07}} & \textbf{68.75{\scriptsize$\pm$0.30}} & \textbf{66.41($\uparrow$1.54)} \\
        \midrule
        0.1\% & LoRA & 87.42{\scriptsize$\pm$1.08} & 66.43{\scriptsize$\pm$0.41} & 69.05{\scriptsize$\pm$4.27} & 70.83{\scriptsize$\pm$0.97} & 73.43 \\
        \rowcolor{LightBlue} & LoRA+TB & \textbf{88.34{\scriptsize$\pm$0.52}} & \textbf{66.79{\scriptsize$\pm$0.73}} & \textbf{69.72{\scriptsize$\pm$3.36}} & \textbf{71.21{\scriptsize$\pm$0.94}} & \textbf{74.02($\uparrow$0.59)} \\
        \midrule
        0.5\% & LoRA & 90.82{\scriptsize$\pm$0.09} & 76.77{\scriptsize$\pm$0.31} & 81.79{\scriptsize$\pm$0.82} & \textbf{78.69{\scriptsize$\pm$0.54}} & 82.02 \\
        \rowcolor{LightBlue} & LoRA+TB & \textbf{91.09{\scriptsize$\pm$0.54}} & \textbf{77.09{\scriptsize$\pm$0.46}} & \textbf{82.02{\scriptsize$\pm$0.41}} & 78.45{\scriptsize$\pm$0.25} & \textbf{82.16($\uparrow$0.14)} \\
        \midrule
        1\% & LoRA & 92.69{\scriptsize$\pm$0.14} & 79.26{\scriptsize$\pm$0.29} & 84.29{\scriptsize$\pm$0.13} & 80.34{\scriptsize$\pm$0.13} & 84.14 \\
        \rowcolor{LightBlue} & LoRA+TB & \textbf{93.04{\scriptsize$\pm$0.10}} & \textbf{79.43{\scriptsize$\pm$0.07}} & \textbf{84.34{\scriptsize$\pm$0.44}}& \textbf{80.51{\scriptsize$\pm$0.16}} & \textbf{84.33($\uparrow$0.19)} \\
        \bottomrule
    \end{tabular}
    }
    \caption{Evaluation results of RoBERTa-base model trained on four larger GLUE tasks. We compare our method (TB) with  Low-Rank Adaptation training (LoRA) fine-tuning. The tasks and their corresponding evaluation metrics are: SST-2 (accuracy), MNLI (accuracy), QNLI (accuracy) and QQP (combined score of F1 score and accuracy)} \label{tb:glue_roberta_lora} 
\end{table*}

\subsection{LoRA Fine-tuning on GLUE}\label{sec:lora_glue_appendix}
\noindent \textbf{Measuring the ESD of LoRA Adapters. }
Some models are too large to fine-tune fully, so one often needs to use LoRA. In this case, LoRA adapters are added to selected layers in the model, and only these adapters are trained during fine-tuning, while the original weight matrix remains fixed.
For a layer with weight matrix $\mathbf{W} \in \mathbb{R}^{d \times k}$ and LoRA adapters $\mathbf{B} \in \mathbb{R}^{d \times r}$ and $\mathbf{A} \in \mathbb{R}^{r \times k}$, we cannot simply calculate ESD of the product between adapters $\mathbf{B} \times \mathbf{A}$, since the rank of the adapters $r \le \min(d, k)$ are low-rank matrices, which would result in a poor ESD landscape. Therefore, for layers with LoRA adapters, we calculate the sum of the product of LoRA adapters and the weight matrix $\mathbf{W}' = \mathbf{W} + \mathbf{B} \times \mathbf{A}$, and then calculate the ESD of its correlation matrix $\mathbf{X} = \mathbf{W}'^\top\mathbf{W}'$.  

We present the results of applying \ourmethod on LoRA Adapters in Table~\ref{tb:glue_roberta_lora}. We can see that \ourmethod consistently achieves higher test results than LoRA alone. We note that our method can at most improve the test accuracy of 3.29\% on 0.02\% SST2 dataset, indicating a significant improvement. From average improvement increases across different tasks, we can see that as we reduce the subsampling ratio, the average improvement of \ourmethod on all tasks continues to increase. This observation aligns with the discussion in Section~\ref{sec:analysis}, that \ourmethod achieves gradually increased gains in fine-tuning performance as the number of tuning data points decreases, further proving the effectiveness of \ourmethod in achieving model alignment in low-data regimes.

\subsection{Question Answering}\label{sec:llama_qa_appendix}
To draw more robust conclusions, we evaluate the empirical performance of \ourmethod on LLM fine-tuning. We choose to fine-tune LLaMA-7B model with LoRA adapters on the ScienceQA dataset~\citep{lu2022learn}. In Table~\ref{tb:scienceqa} we report the test accuracy of LoRA and \ourmethod under different subsampling ratios on ScienceQA dataset. We can see that \ourmethod continues to yield better test performance on low-data regimes.

\begin{table}[!h]
    \centering
    \resizebox{0.45\textwidth}{!}{
    \begin{tabular}{c|ccc}
        \toprule
        \bf{Subsampling} & & & \\
        \bf{Ratio} & 1\% & 5\% & 10\% \\
        \midrule
        LoRA & 51.12{\scriptsize$\pm$0.87} & 65.24{\scriptsize$\pm$1.04} &  73.40{\scriptsize$\pm$0.39} \\
        \midrule
        \rowcolor{LightBlue}LoRA+TB & \textbf{53.09{\scriptsize$\pm$1.64}} & \textbf{65.96{\scriptsize$\pm$1.21}} & \textbf{73.70{\scriptsize$\pm$0.80}} \\
        \bottomrule
    \end{tabular}
    }
    \caption{Test accuracy (\%) on ScienceQA dataset of LLaMA-7B model trained with different subsampled training set.} \label{tb:scienceqa} 
\end{table}

\subsection{Training FNO and UNet Model on DarcyFlow Dataset}\label{sec:darcy_appendix}
In Table~\ref{tb:darcyflow_fno_unet} we show the test results of training the FNO and UNet model on the DarcyFLow dataset with a subsampling ratio ranging from 0.6\% to 100\%, evaluated by Normalized Root Mean Squared Error (nRMSE). We show that \ourmethod achieves lower nRMSE compared to the baseline on all subsampling ratios. Specifically, \ourmethod reduces the nRMSE of the UNet model trained on 2.5\% of the DarcyFlow dataset by a significant 10.89\%, and improve the nRMSE of FNO on 0.6\% by 9.71\%. 

\begin{table}[!h]
    \centering
    \resizebox{0.45\textwidth}{!}{
    \begin{tabular}{c|c|cc}
        \toprule
        \bf{Subsampling} &  &  &  \\
        \bf{Ratio} & \bf{Method} & \bf{FNO} & \bf{UNet} \\
        \midrule
        & Baseline & 2.58e-03{\scriptsize$\pm$2.69e-05} & 5.27e-03{\scriptsize$\pm$3.27e-05} \\
        \rowcolor{LightBlue} 
        100\% & TB & \textbf{2.52e-03{\scriptsize$\pm$5.68e-05}} & \textbf{5.07e-03{\scriptsize$\pm$1.41e-05}} \\
        & Error Reduced & 2.33\% & 3.80\% \\
        \midrule
        
        & Baseline & 1.04e-02{\scriptsize$\pm$4.11e-04} & 1.43e-02{\scriptsize$\pm$1.21e-03} \\
        \rowcolor{LightBlue} 10.0\% & TB & \textbf{1.01e-02{\scriptsize$\pm$1.30e-04}} & \textbf{1.34e-02{\scriptsize$\pm$9.50e-04}} \\
        & Error Reduced & 2.88\% & 6.29\% \\
        \midrule
        
        & Baseline & 1.76e-02{\scriptsize$\pm$5.17e-04} & 1.98e-02{\scriptsize$\pm$1.79e-03} \\
        \rowcolor{LightBlue} 5.0\% & TB & \textbf{1.62e-02{\scriptsize$\pm$2.19e-04}} & \textbf{1.81e-02{\scriptsize$\pm$1.35e-03}} \\
        & Error Reduced & 7.95\% & 8.59\% \\
        \midrule
        
        & Baseline & 2.88e-02{\scriptsize$\pm$9.79e-04} & 2.57e-02{\scriptsize$\pm$9.89e-04} \\
        \rowcolor{LightBlue} 2.5\% & TB  & \textbf{2.64e-02{\scriptsize$\pm$5.72e-04}} & \textbf{2.29e-02{\scriptsize$\pm$1.94e-03}} \\
        & Error Reduced & 8.33\% & 10.89\% \\
        \midrule
        
        & Baseline & 6.28e-02{\scriptsize$\pm$1.78e-03} & 4.59e-02{\scriptsize$\pm$3.10e-03} \\
        \rowcolor{LightBlue} 0.6\% & TB & \textbf{5.67e-02{\scriptsize$\pm$1.62e-03}} & \textbf{4.45e-02{\scriptsize$\pm$1.48e-03}} \\
        & Error Reduced & 9.71\% & 3.05\% \\
        \bottomrule
    \end{tabular}
    }
    \caption{Evaluation results of FNO and UNet model trained on DarcyFlow ($\beta=100$) dataset. We compare our method (TB) with the baseline. The evaluation metric is nRMSE ($\downarrow$).
    } \label{tb:darcyflow_fno_unet}
\end{table}

\section{Compute Resources}
We conduct our experiments on Quadro RTX 6000, NVIDIA L40(40GB), and NVIDIA RTX A6000 GPU clusters. Specifically, we run every full fine-tuning of RoBERTa-base on GLUE and SuperGLUE datasets using one Quadro RTX 6000 GPU per job. For each of the LoRA fine-tuning of RoBERTa-base on GLUE tasks, we utilize a single NVIDIA RTX A6000 GPU to train the model. For LLaMA-7B LoRA fine-tuning experiments, we use 4 NVIDIA RTX A6000 GPUs for one job. For all Neural PDE experiments, we use a single NVIDIA L40(40GB) GPU for each job.

\section{More Ablation Study Results}\label{appendix:ablation}

\subsection{Different ESD metrics and scheduling functions in using \ourmethod in SciML.}\label{appendix:ablation_sciml}
We compare the performance of using different ESD measuring metrics and scheduling functions of \ourmethod on SciML tasks. Table~\ref{tb:tb_func_comparison_sciml} reports the results of different \ourmethod settings in training the FNO model on solving the 1DCFD task. We can see that \ourmethod outperforms the baseline method at every subsampling ratio, and our proposed scaling function \SIGMOID achieves more stable performance than \LINEARMAP. At most subsampling ratios, using \ALPHAHILL we can achieve results that are comparable to or even better than those obtained with other metrics.

\begin{table*}[!h]    
    \centering
    \resizebox{0.95\textwidth}{!}{
    \begin{tabular}{c|cccccc}
        \toprule
        \bf{Ratio} & 100\% & 50.0\% & 25.0\% & 10.0\% & 2.5\% & 0.6\%\\
        \midrule
        Baseline & 5.02e-02{\scriptsize$\pm$4.43e-03} & 6.04e-02{\scriptsize$\pm$3.17e-03} & 
        7.81e-02{\scriptsize$\pm$3.79e-03} &
        1.13e-01{\scriptsize$\pm$4.79e-03} &
        2.11e-01{\scriptsize$\pm$2.79e-03} &
        2.48e-01{\scriptsize$\pm$3.35e-03} \\
        \midrule
        
        \LINEARMAP & 4.95e-02{\scriptsize$\pm$3.49e-03} & 5.70e-02{\scriptsize$\pm$5.52e-04} & 
        \textbf{7.26e-02{\scriptsize$\pm$1.02e-03}} &
        1.02e-01{\scriptsize$\pm$3.00e-03} & 
        2.05e-01{\scriptsize$\pm$4.77e-03} & 
        2.40e-01{\scriptsize$\pm$7.47e-03} \\
        \midrule
        
        \SIGMOID(\ALPHAHILL) & \textbf{4.74e-02{\scriptsize$\pm$6.57e-04}} & 
        \textbf{5.68e-02{\scriptsize$\pm$2.28e-03}} & 
        7.42e-02{\scriptsize$\pm$1.87e-03} &
        \textbf{1.02e-01{\scriptsize$\pm$1.88e-03}} & 
        2.08e-01{\scriptsize$\pm$5.25e-03} & 
        2.38e-01{\scriptsize$\pm$2.84e-03} \\
        \midrule
         
        \SIGMOID(\STABLERANK) & 4.89e-02{\scriptsize$\pm$2.03e-03} & 6.03e-02{\scriptsize$\pm$7.47e-04} & 
        7.32e-02{\scriptsize$\pm$1.73e-03} &
        1.06e-01{\scriptsize$\pm$4.85e-03} & 
        2.07e-01{\scriptsize$\pm$1.36e-03} & 
        2.45e-01{\scriptsize$\pm$6.11e-03} \\
        \midrule
        
        \SIGMOID(\SPECTRALNORM) & 4.84e-02{\scriptsize$\pm$2.86e-03} & 
        5.77e-02{\scriptsize$\pm$1.48e-03} & 
        7.50e-02{\scriptsize$\pm$5.70e-03} &
        1.03e-01{\scriptsize$\pm$4.66e-03} & 
        \textbf{1.91e-01{\scriptsize$\pm$1.05e-02}} & 
        \textbf{2.34e-01{\scriptsize$\pm$1.12e-03}} \\
        \bottomrule
    \end{tabular}
    }
    \caption{Comparing different Temperature Balancing scheduling algorithm and ESD metrics on FNO model trained with 1DCFD dataset. The \ourmethod series functions can help models achieve lower test nRMSE among all subsampling ratios, and the \SIGMOID outperform the original \LINEARMAP function.} \label{tb:tb_func_comparison_sciml}
\end{table*}

\section{More Analysis Results}\label{appendix:analysis}
\subsection{Diagnosing the Data Limitation Using HT Metrics}\label{appendix:diagnosis}
\begin{figure*}[!ht]
    \centering
    \begin{subfigure}{0.45\linewidth}
        \includegraphics[width=\textwidth]{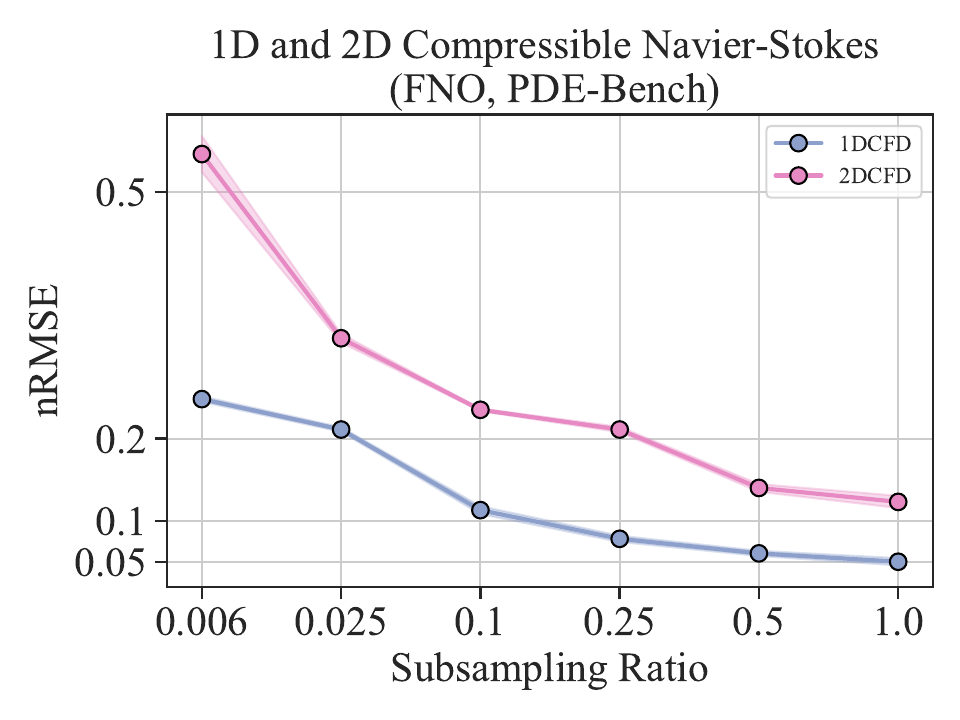}
        \caption{nRMSE $(\downarrow)$}
        \label{fig:nrmse_trend_fno_cfd}
    \end{subfigure}
    \hfill
    \centering
    \begin{subfigure}{0.45\linewidth}
        \includegraphics[width=\textwidth]{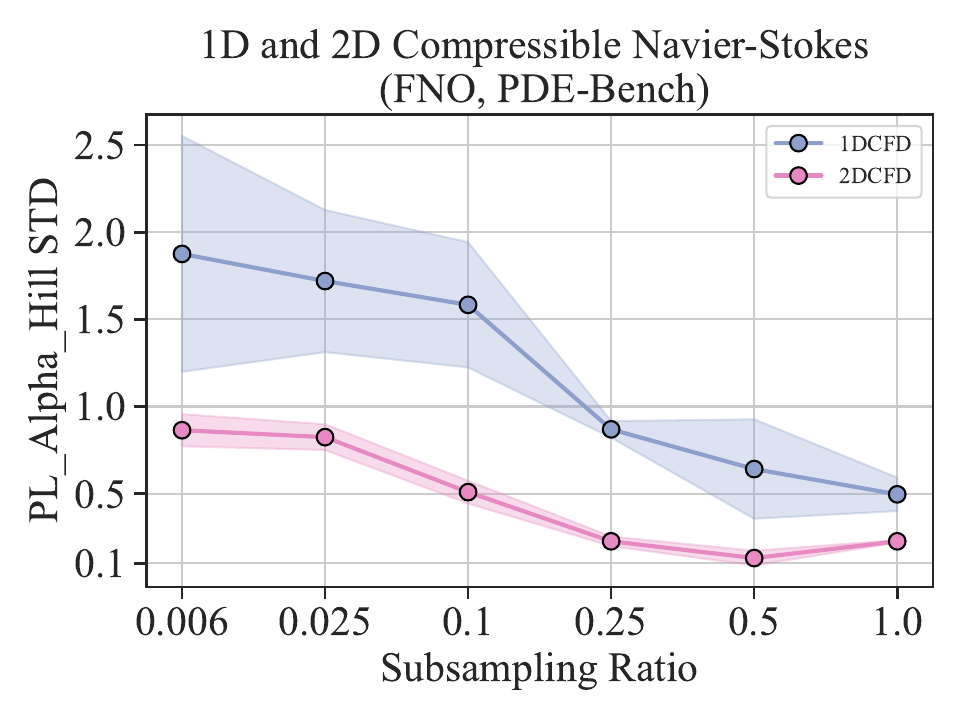}
        \caption{STD of layer-wise \ALPHAHILL}
        \label{fig:alpha_std_trend_fno_cfd}
    \end{subfigure}
    \caption{Predicting model performance under different training data using the variance of layer-wise \ALPHAHILL. ~\ref{fig:nrmse_trend_fno_cfd} shows the trend of test performance of FNO model on 1D and 2D CFD datasets. ~\ref{fig:alpha_std_trend_fno_cfd} shows the trend of standard deviation of \ALPHAHILL across different FNO layers in different training data.}
    \label{fig:trend_plot_fno_cfd}
\end{figure*}

Following Section~\ref{sec:diagnosis}, here we further analyzed FNO model's test performance using Alpha-related metrics as the training data size decreases. Figure~\ref{fig:trend_plot_fno_cfd} demonstrates that the change of the STD of \ALPHAHILL corresponds very closely with the variations in the model's performance. We observe that as the subsampling ratio decreases, the nRMSE on the 1D and 2D CFD PDEs solving increases, indicating a deterioration in model's performance. Simultaneously, the STD of \ALPHAHILL becomes larger, suggesting that the training across the model layers is becoming increasingly uneven. Therefore, the STD of \ALPHAHILL effectively captures the model's performance variations in response to changes in the amount of training data, which aligns closely with the results obtained in our previous experiments in Figure~\ref{fig:alpha_tb_std_trend}.

\subsection{More Analysis Study Results in the STD of \ALPHAHILL}\label{appendix:analysis_alpha_std}
\begin{figure*}[!h]
    \centering
    \begin{subfigure}{0.45\linewidth}
        \includegraphics[width=\textwidth]{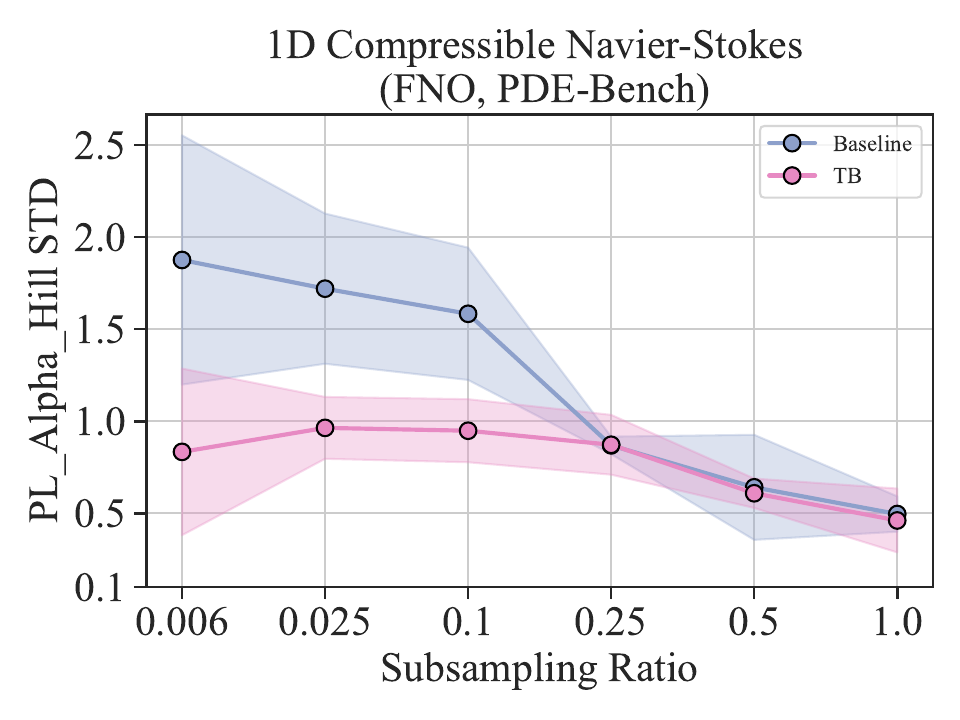}
        \caption{STD of layer-wise \ALPHAHILL in training FNO on 1DCFD}
        \label{fig:alpha_std_fno_1dcfd}
    \end{subfigure}
    \hfill
    \centering
    \begin{subfigure}{0.45\linewidth}
        \includegraphics[width=\textwidth]{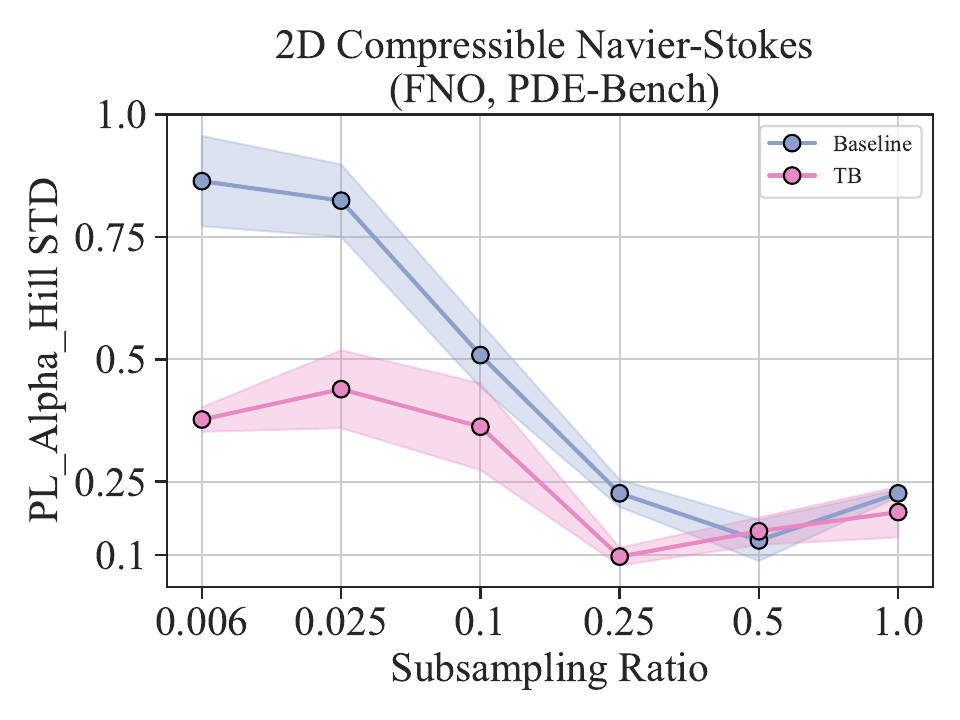}
        \caption{STD of layer-wise \ALPHAHILL in training FNO on 2DCFD}
        \label{fig:alpha_std_fno_2dcfd}
    \end{subfigure}
    \caption{Comparing the STD of layer-wise \ALPHAHILL measured in using baseline method and \ourmethod training FNO model on 1D and 2D CFD datasets. The results demonstrate that \ourmethod can reduce the STD, and this effect is more significant when the subsampling ratio is smaller, indicating that our approach helps ensure more uniform training across each layer of the model.}
    \label{fig:alpha_std_fno_cfds}
\end{figure*}

In Figure~\ref{fig:alpha_tb_std_trend} and~\ref{fig:alpha_std_fno_cfds}, we compare the STD of the \ALPHAHILL between the baseline and \ourmethod on fine-tuned LLM and trained FNO models at different subsampling ratios. When the subsampling ratio is relatively large, the STD of \ALPHAHILL of models is smaller, and the impact of the \ourmethod method on this metric is also minimal. However, when the subsampling ratio is relatively small, the opposite is true: the \ourmethod method makes the distribution of \ALPHAHILL across each layer of the model more uniform.

\begin{figure}[!h]
    \centering
    \includegraphics[width=0.5\linewidth]{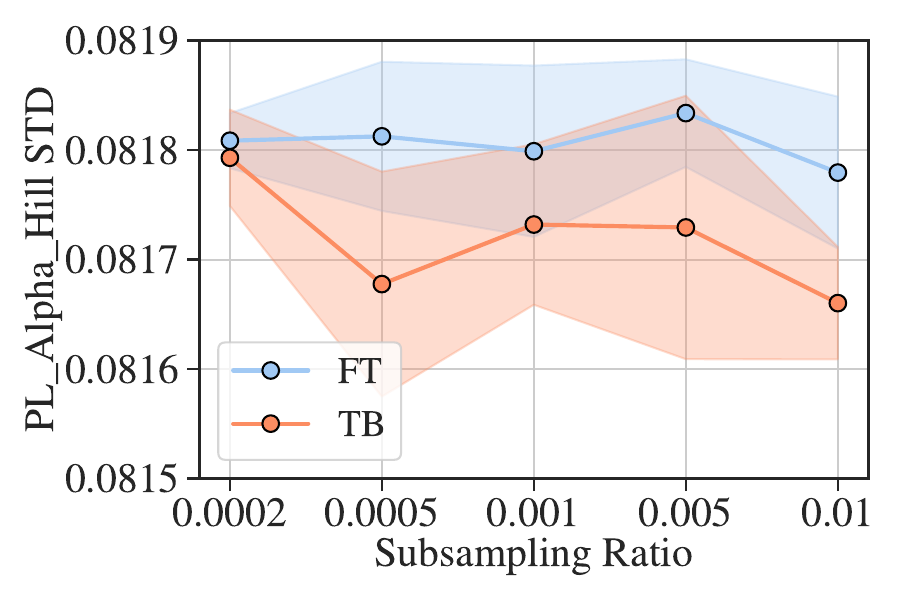}
    \caption{Analyzing the distribution of \ALPHAHILL of baseline FT and \ourmethod on RoBERTa-base model trained on QNLI across different subsampling ratios. We observe that \ourmethod continues to show lower STD of \ALPHAHILL, suggesting a more evenly distributed \ALPHAHILL. 
    } 
    \label{fig:alpha_tb_std_trend}

\end{figure}

\end{document}